%% file: main.tex
\documentclass[a4paper, 11pt]{article}
\pdfoutput=1
\usepackage{etoolbox}

\input{preamble/_preamble_includes}

\title{When Do Discrete Actions Make Good Continuous Controllers?}

\author{
Zeji Yi$^{1,2,\dagger}$, Soham Dasgupta$^{1,2,\dagger}$, Xinyang Zhang$^{2}$, Koushil Sreenath$^{2}$, Guannan Qu$^{1}$, \\
Xi Chen$^{2}$, Max Simchowitz$^{1,2,*}$, Guanya Shi$^{1,2,*}$ \\
\vspace{-.3em}
\rule{.38\textwidth}{.7pt}
\\
\footnotesize
$^{1}$Carnegie Mellon University \quad
$^{2}$Amazon FAR (Frontier AI \& Robotics) \\
\footnotesize
$^{\dagger}$Work done during internship at Amazon FAR. \quad $^*$Equal advising.
}

\date{\today}
\paperwebsite{https://iscoyizj.github.io/discrete-action-website/}

\begin{document}

\begin{tcolorbox}[
    colback=steelgray, colframe=irongray,
    boxrule=0pt,
    arc=2mm
  ]
  \maketitle
  \tcbline
  \input{body/teaser_fig}\par
  \tcbline
  \ifdefempty{\metadatalist}{}{\metadatalist\par}
\end{tcolorbox}
\thispagestyle{plain}

\begin{abstract}
  \input{body/abstract}
\end{abstract}
\input{body/intro}
\input{body/prelim}
\input{body/discrete_vs_continuous}
\input{body/factorization_error}
\input{body/geometry_and_tokenization}
\input{body/discussion}
\input{body/statements}

\newpage
\bibliographystyle{plainnat}
\bibliography{refs}

\newpage
\tableofcontents

\newpage
\appendix
\crefalias{section}{appendix}
\crefalias{subsection}{appendix}
\crefalias{subsubsection}{appendix}
\input{appendix/related_work}
\FloatBarrier
\Needspace{10\baselineskip}
\input{appendix/environments}
\FloatBarrier
\Needspace{10\baselineskip}
\input{appendix/action_heads}
\Needspace{10\baselineskip}
\input{appendix/results}
\FloatBarrier
\Needspace{10\baselineskip}
\input{appendix/responsiveness}
\FloatBarrier
\Needspace{10\baselineskip}
\input{appendix/factorization_error}
\FloatBarrier
\Needspace{10\baselineskip}
\input{appendix/tokenization_and_geometry}
\FloatBarrier
\Needspace{10\baselineskip}
\input{appendix/synthetic}

\end{document}

%% file: preamble/_preamble_includes.tex
\input{preamble/color_defs}
\input{preamble/header}
\input{preamble/characters}
\input{preamble/general_macros}
\input{preamble/specific_macros}

\input{preamble/arxiv_title}

%% file: preamble/color_defs.tex
\usepackage{xcolor}
\usepackage{tcolorbox}
\tcbuselibrary{skins,breakable}

\definecolor{contrarydarkcolor}{HTML}{006666}
\definecolor{irongray}{HTML}{6D6E71}
\definecolor{steelgray}{HTML}{F1F4F7}
\newcommand{\linkcolor}{contrarydarkcolor}
\newcommand{\urlcolor}{contrarydarkcolor}
\newcommand{\citecolor}{contrarydarkcolor}

%% file: preamble/header.tex
\usepackage[T1]{fontenc}

\usepackage{natbib}
\usepackage{amssymb,bm}
\usepackage{mathtools}
\usepackage[english]{babel}
\usepackage{booktabs}
\usepackage[shortlabels]{enumitem}
\usepackage{subcaption}
\usepackage{graphicx,tabularx,array,wrapfig,placeins,needspace}
\usepackage[table]{xcolor}
\usepackage{microtype}

\usepackage[colorlinks=true,linkcolor=\linkcolor,urlcolor=\urlcolor,citecolor=\citecolor,breaklinks]{hyperref}
\usepackage[capitalise,nameinlink]{cleveref}

\usepackage[
  letterpaper,
  left=1in,
  right=1in,
  top=0.9in,
  bottom=0.9in,
  headheight=14pt,
  headsep=14pt,
  footskip=20pt
]{geometry}

\usepackage{fancyhdr}
\usepackage[bitstream-charter]{mathdesign}

\usepackage[scaled=0.92]{PTSans}

\numberwithin{equation}{section}

%% file: preamble/characters.tex
\newcommand{\bzero}{\ensuremath{\mathbf 0}}
\def\ba{\ensuremath{\mathbf{a}}}
\def\bp{\ensuremath{\mathbf{p}}}
\def\bq{\ensuremath{\mathbf{q}}}
\def\bv{\ensuremath{\mathbf{v}}}
\def\bx{\ensuremath{\mathbf{x}}}
\def\bz{\ensuremath{\mathbf{z}}}
\def\cD{\ensuremath{\mathcal{D}}}
\def\cL{\ensuremath{\mathcal{L}}}

%% file: preamble/general_macros.tex
\newcommand{\E}{\mathbb{E}}
\newcommand{\R}{\mathbb{R}}
\newcommand{\eye}{\mathbf{I}}

\makeatletter
\newcommand\addtometadatalist[5][]{%
    \begingroup
    \if\relax#3\relax\def\sep{}\else\def\sep{#5}\fi
    \let\protect\@unexpandable@protect
    \xdef#3{\expandafter{#3}\sep #4[#1]{#2}}%
    \endgroup
}
\newcommand\metadatalist{}
\newcommand\metadataformat[2][]{{\small \textbf{#1:} #2}}
\newcommand\metadata[2][]{\addtometadatalist[#1]{#2}{\metadatalist}{\metadataformat}{\\}}
\makeatother

\newcommand{\paperwebsite}[1]{\metadata[Website]{\url{#1}}}
\renewcommand\date[1]{\metadata[Date]{#1}}

%% file: preamble/specific_macros.tex
\Crefname{equation}{Eq.}{Eqs.}
\Crefname{figure}{Fig.}{Figs.}
\Crefname{table}{Tab.}{Tabs.}

\colorlet{QuestionAccent}{contrarydarkcolor}
\definecolor{HypothesisAccent}{HTML}{774A80}
\newcommand{\papercallout}[3]{%
  \par\addvspace{.7\baselineskip}\begingroup
  \setlength{\fboxsep}{7pt}\setlength{\fboxrule}{.6pt}%
  \noindent\fcolorbox{#1!65!white}{#1!4!white}{%
    \parbox{\dimexpr\linewidth-2\fboxsep-2\fboxrule\relax}{%
      \textcolor{#1}{\textbf{#2}}\enspace #3}}%
  \par\endgroup\addvspace{.7\baselineskip}}
\newcommand{\questionbox}[1]{\papercallout{QuestionAccent}{Question}{\textbf{#1}}}
\newcommand{\takeaway}[1]{\papercallout{QuestionAccent}{Takeaway}{#1}}
\newcommand{\hypothesisheading}[1]{\paragraph{\textcolor{HypothesisAccent}{#1}}}

\newcommand{\confpar}[1]{\paragraph{#1}}

\tcbset{
  shortbox/.style={
      width=\linewidth,
      top=0pt,
      bottom=0pt,
      left=1.5pt,
      right=1.5pt,
      colback=blue!60!gray!5,
      colframe=black,
      colbacktitle=black,
      enhanced,
      center,
      attach boxed title to top left={yshift=-0.1in,xshift=0.15in},
      boxed title style={boxrule=0pt,colframe=white,},
    }
}
\newtcolorbox{ssbox}[2][]{shortbox,title=#2,#1}

\newcommand{\beps}{\bm{\epsilon}}

%% file: body/teaser_fig.tex
\begin{minipage}[t]{1.0\linewidth}
    \captionsetup{type=figure}
    \centering
    \includegraphics[width=\linewidth]{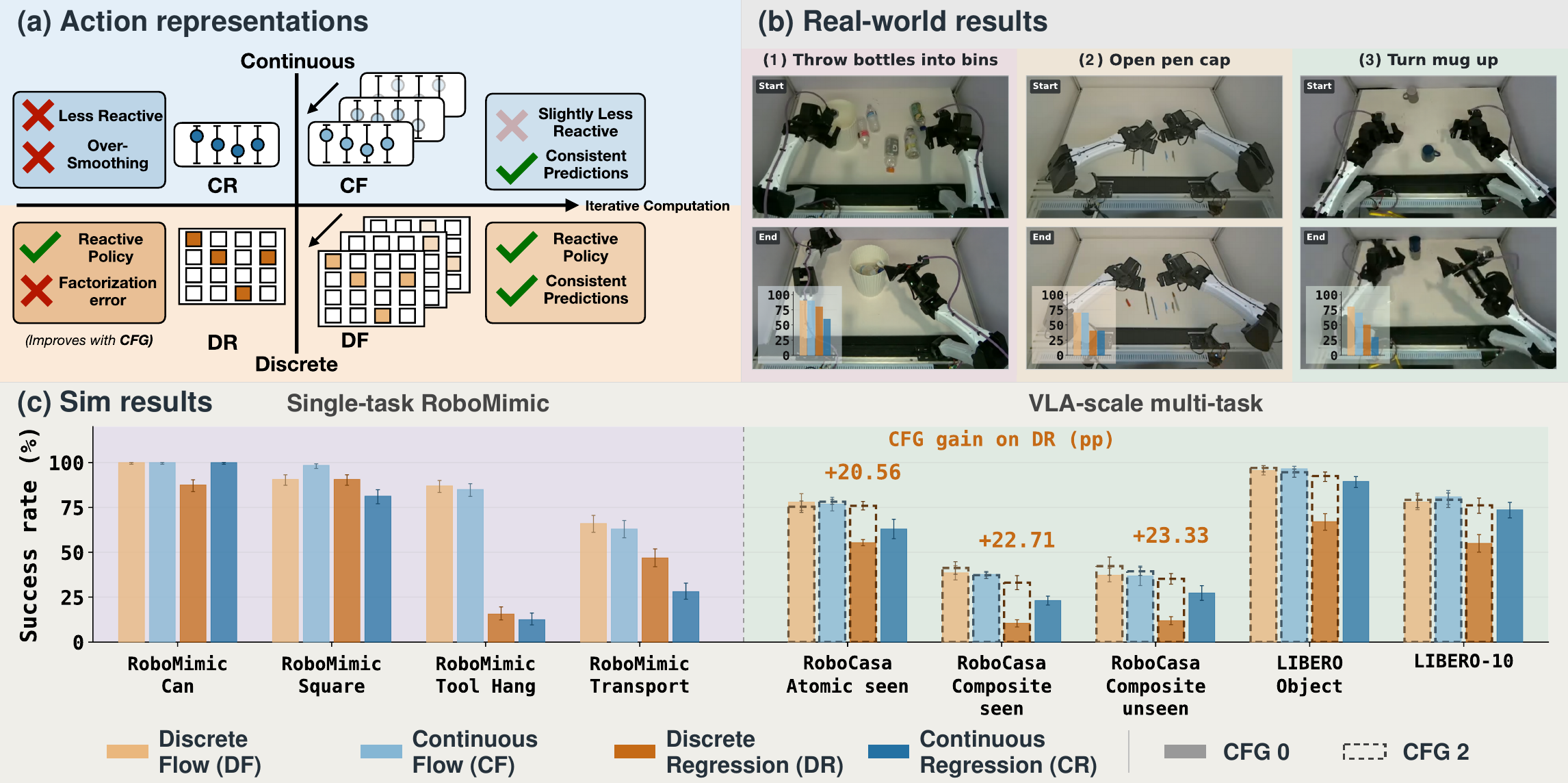}
    \caption{\textbf{(a)} Action Representation and comparison; \textbf{(b)} VLA-real (three real-world tasks); \textbf{(c)} Closed-loop success on RoboMimic and
    VLA-sim (RoboCasa and LIBERO);
    Within each group, heads appear in the order DF, CF, DR, CR; orange denotes discrete heads and blue denotes continuous heads. For RoboCasa and LIBERO DF, CF, and DR, filled bars show CFG~0 and transparent dashed bars show CFG~2.
      The heads are described in \Cref{app:action-heads}, the evaluation protocols in \Cref{app:environments}, and full results in \Cref{app:results}; RoboMimic bars report each head family's best configuration (\Cref{app:results-robomimic}).}
    \label{fig:robot-success}
\end{minipage}

%% file: body/abstract.tex
Though modern AI models like LLMs  predict discrete tokens, today's robotic foundation models represent actions as continuous vectors. In this paper, we challenge this common consensus by asking if continuous action parametrizations are superior to their discrete counterparts.  We conduct an extensive comparison of single-step regression and multi-step flow variants of both continuous and discrete policies in both from-scratch single-task and multi-task VLA training, in both simulation and on real hardware. In all regimes, we
find that discrete action prediction, even the most naive, uniform-bin quantization,  can match or even outperform continuous-action policies in task success rate. Moreover, we find that the challenge to discrete action policies is \emph{not} action precision, but rather factorization error introduced by product-distribution predictions across time and action dimensions. We show that, when paired with the correct interventions---iterative computation through discrete flows, or distribution sharpening through classifier-free guidance---discrete policies can turn this weakness into a strength, enabling more responsive action changes during a predicted action chunk. Lastly, we show that attempts to explicitly inject geometric information into discrete policies, e.g. through corruption processes that emulate continuous flow matching, do \emph{not} yield consistent improvements. Taken together, our findings question existing assumptions that robotic action prediction requires distinct design decisions compared to other AI domains, and motivates the possibility of discrete action prediction as a compelling alternative to today's continuous action parametrizations. Link for the webpage \href{https://iscoyizj.github.io/discrete-action-website/}{https://iscoyizj.github.io/discrete-action-website/}

%% file: body/intro.tex
\section{Introduction}

A major goal of modern robot learning is to perform complex and varied tasks to high precision. As a result, the community has devoted significant study to the question of how to parametrize the actions predicted by robotic policies and foundation models, such as vision-language-action models (VLAs)~\citep{brohan2023rt,Kim2024OpenVLAAO,black2024pi0,nvidia2025gr00t} and world action models (WAMs)~\citep{li2025uva,ye2026worldaction,kim2026cosmospolicy,li2026causalworld}.  The common consensus is that robot models should predict continuous-valued actions during deployment, as these are lossless and therefore support high-precision behavior. For example, most of today's VLAs predict actions using continuous-space generative models, such as flow~\citep{black2024pi0,physicalintelligence2025pi05} or diffusion models~\citep{chi2023diffusion}, which are believed to better capture training data diversity whilst enhancing the precision of action predictions.

At the same time, modern robot learning is adopting increasingly more features from other AI domains, notably language modeling. These include test-time reasoning \citep{wei2022chain}, on-task self-correction \citep{huang2022inner}, future goal prediction \citep{du2023learning}, and long-term memory \citep{Wang2023VoyagerAO}. In fact, recent work has showcased the possibility that frontier language models can directly control robot embodiments through text generation \citep{brohan2023rt, Kim2024OpenVLAAO}. Yet, today's frontier AI models, even ones that process multiple modalities such as text, images, and audio, both take as input and produce discretely quantized tokens. Whether robot actions should follow the same discrete format therefore raises a key question:

 \begin{ssbox}{}
Are robot actions unique among other AI modalities in that they must be predicted as continuous values, or can \textbf{discrete action representations} compete with, or even outperform, their continuous counterparts?
\end{ssbox}

\subsection{Contributions}
In this work, we undertake a systematic comparison between the performance of continuous and discrete action parameterizations for both single-task robot policies and multi-task vision-language-action (VLA) models. In short, we show that \textbf{discrete action parametrization can be competitive with, and even uniformly outperform, continuous action prediction}, provided that correct design decisions are made. These findings provide a deeper understanding of the performance of recent discrete diffusion/flow VLA policies \citep{liang2025discrete,wen2025llada,chen2025udvla}.

In greater detail, we consider parametrizations of the \emph{action head}, or part of a robot model responsible for action prediction. Our comparison builds on classical supervised control and imitation learning~\citep{pomerleau1988alvinn,ross2011reduction,osa2018algorithmic}. We focus our study on 4 representative parametrizations: (1) continuous regression (CR), which predicts continuous-valued actions in a single forward pass and is trained with a regression objective~\citep{bojarski2016end,kim2025oft} (2) continuous flow (CF), which generates continuous actions using flow matching, a family applied to robot control and motion planning~\citep{hu2024adaflow,chisari2024pointflowmatch,black2024pi0,zhang2024flowpolicy,nguyen2025flowmp} (3) discrete regression (DR), which predicts a quantized action via cross-entropy classification~\citep{brohan2022rt1} and (4) discrete flow (DF), which predicts quantized actions in multiple steps via the discrete flow matching objective~\citep{gat2024discrete}. Hybrid methods such as Behavior Transformer~\citep{shafiullah2022behavior} combine discrete action-bin prediction with a continuous offset; in contrast, we study purely discrete and continuous action heads separately to isolate the effect of action representation. Recent work compares regression and flow policies under architecture-matched settings~\citep{pan2025noising}. We evaluate these parametrizations from single-task policies on RoboMimic~\citep{robomimic2021} to multi-task VLA-scale policies using RoboCasa~\citep{robocasa365}, DROID~\citep{khazatsky2024droid}, and LIBERO~\citep{liu2023libero}, as well as real-world robot experiments. All are evaluated using closed-loop task success, action prediction accuracy, and within-chunk temporal variation.
We identify 3 key findings that upend prior beliefs about the relative performance of discrete and continuous action modalities, challenging the prevailing view that precise robot control requires predicting continuous action prediction:

\begin{itemize}[leftmargin=8pt]
    \item \textbf{Finding 1: Control Precision}  Contrary to prior belief, discrete action parametrizations, even discrete regression (DR) with the simplest bin discretization, can exhibit greater control precision than their continuous counterparts, despite the possibly lossy quantization of actions! We support this claim in both sim and real settings in \Cref{sec:discrete-vs-continuous}. We show that this arises from greater policy reactivity, both to the action-chunk time index and the conditioning observation.
    \item \textbf{Finding 2: Iterative Refinement and Guidance Mitigate Factorization Error} (\Cref{sec:factorization}). Independently predicted tokens can combine incompatible choices across time and action dimensions, even when individual token distributions are accurate. DF solves this by iterative computation, while classifier-free guidance reduces ambiguity by changing token preferences. Both improve control significantly, supporting joint consistency as main explanation for their gains. In the RoboCasa comparison, guidance raises DR success from 27.1\% to 49.2\%, approaching DF's 53.9\% and CF's 52.7\%, largely ahead of CR's 38.8\%.
    \item \textbf{Finding 3: Explicit Geometry and Structured Tokenization Provide Limited Additional Gains} (\Cref{sec:geometry-and-tokenization}).
    Encoding action distances or using alternative tokenizers improves some configurations, but none exceeds our best DF result with uniform bins and the standard mixture path. Adjacent bins also acquire similar output embeddings, suggesting that discrete heads can learn local geometry without encoding it explicitly.
\end{itemize}
We discuss related work in \Cref{app:related}.

%% file: body/prelim.tex
\section{Preliminaries: Action Representations and Prediction Heads}
\label{sec:prelim}

\confpar{Notation.}
Let $o$ denote the policy's conditioning information, such as images, language, and proprioception, and let $\ba_{1:H}=(\ba_1,\ldots,\ba_H)\in\R^{H\times M}$ be an action chunk ~\citep{zhao2023act} of horizon $H$, with $M$ action coordinates per timestep. We represent actions in \emph{end-effector (EEF) space}, with 3 translation coordinates $\ba^{\mathrm{tr}}\in\R^3$, rotation coordinates $\ba^{\mathrm{rot}}$, and a binary gripper coordinate $\ba^{\mathrm{g}}\in\{0,1\}$, and $\ba_i = [\ba^{\mathrm{tr}}_i,\ba^{\mathrm{rot}}_i,\ba^{\mathrm{g}}_i]$. All policies use \emph{relative} EEF actions: every target in the chunk is measured from the EEF pose at prediction time. The VLA policies encode the rotation as an axis-angle vector $\ba^{\mathrm{rot}}\in\R^3$, and the single-task RoboMimic policies as a 6D representation $\ba^{\mathrm{rot}}\in\R^6$ of a rotation matrix in $\mathrm{SO}(3)$. \Cref{app:results-action-space} compares relative EEF actions with delta and absolute EEF actions and with relative joint-space actions.  From training pairs $(o,\ba_{1:H})\sim\cD$, we learn
 (possibly stochastic) policies $\hat\ba_{1:H} \sim \pi_\theta(o)$ that execute the chunk open
loop and replan.
All heads share a backbone design (\Cref{app:head-arch}); full
objectives and implementation details are in
\Cref{app:action-heads}.

\confpar{Continuous regression (CR).}
The policy $\pi_\theta^{\mathrm{CR}}(o)$ predicts a chunk in one pass
using squared-error regression. Its population minimizer is the
conditional mean $\E[\ba_{1:H}\mid o]$, which need not be a demonstrated
action when several behaviors are plausible.

\confpar{Continuous flow (CF).}
Flow matching learns a conditional velocity field
$v_\theta$~\citep{lipman2023flow,liu2022rectified}.
From $\bx_0=\beps_{\mathrm{C}}\sim\mathcal N(0,\eye)$, generation follows
$\dot\bx_t=v_\theta(\bx_t,t,o)$ over $t\in[0,1]$.
After $N$ Euler integration steps, the policy returns
$\hat \ba_{1:H} \gets \bx_{t_N}$,  with $t_N=1$. Here,
flow time $t$ indexes generation, rather than the action horizon.
The interpolant and training objective are given in
\Cref{app:head-continuous}.

\confpar{Discrete action representation.}
We quantize each action coordinate into one of $B$ bins after
normalization and clipping. Let $[B]=\{1,\ldots,B\}$ and $L=HM$.
Flattening the quantized chunk in a fixed order gives
$\bz=(z_1,\ldots,z_L)\in[B]^L$, whose entries we call \emph{tokens}.
Thus, $B$ is both the number of bins per coordinate and the action-token
vocabulary size; unless otherwise stated, $B=512$.
The quantizer $Q:\R^{H\times M}\to[B]^L$ and decoder
$D_Q:[B]^L\to\R^{H\times M}$ satisfy
$\bz=Q(\ba_{1:H})$ and $\hat\ba_{1:H}=D_Q(\hat\bz)$.
{}
We primarily use \textbf{uniform quantization}. For action coordinate
$i$, let $[\ell_i,u_i]$ be its range and
$\Delta_i=(u_i-\ell_i)/B$ its bin width. With scalar action $a_{h,i}$
and token position $j=(h-1)M+i$, encoding and decoding are $
 Q(\ba_{1:H})_j:=\operatorname{clip}_{[1,B]}
 \left(1+\left\lfloor\frac{a_{h,i}-\ell_i}{\Delta_i}\right\rfloor\right)$,$
 [D_Q(\bz)]_{h,i}
 :=\ell_i+(z_j-\tfrac12)\Delta_i.$
Clipping keeps the bin index in $[B]$, including at the upper endpoint;
decoding returns the bin center and restores the chunk shape.
Both maps use normalized coordinates, with normalization inverted
before execution. Recipe details and alternative tokenizations are
described in \Cref{app:head-quantization}.

\confpar{Discrete regression (DR).}
In one forward pass, the head outputs $L\times B$ logits whose
position-wise softmax defines
$p_{\theta,j}^{\mathrm{DR}}(\cdot\mid o)$~\citep{brohan2022rt1}.
Conditioned on the model inputs, it factorizes the token distribution as
$p_\theta^{\mathrm{DR}}(\bz\mid o)=\prod_{j=1}^L
p_{\theta,j}^{\mathrm{DR}}(z_j\mid o)$.
For $\bz=Q(\ba_{1:H})$, we train with tokenwise cross-entropy:
$
 \cL_{\mathrm{DR}}
 =\E_{(o,\ba_{1:H})\sim\cD}
 \left[{\textstyle-\frac{1}{L} \sum_{j=1}^L}
 \log p_{\theta,j}^{\mathrm{DR}}(z_j\mid o)\right].
$
The factorization makes maximizing the model's joint probability
tractable: we greedily select
$\hat z_j=\arg\max_{k\in[B]}p_{\theta,j}^{\mathrm{DR}}(k\mid o)$
at each position~\citep{Devlin2019BERTPO}. For the VLA policies, DR is
trained with a distance-smoothed version of this loss
(\Cref{app:head-objectives}).
Writing $\hat\bz=(\hat z_1,\ldots,\hat z_L)$, the policy returns
$\hat \ba_{1:H}:=D_Q(\hat\bz)$.

\confpar{Discrete flow (DF).}
Discrete flow~\citep{gat2024discrete} generates a token chunk through
iterative updates, starting from a product source
$q_0(\bz_0)=\prod_{j=1}^L q_{0,j}(z_{0,j})$.
Let $\alpha:[0,1]\to[0,1]$ be a continuously differentiable,
nondecreasing schedule with $\alpha(0)=0$, $\alpha(1)=1$, and
$\alpha(t)<1$ for $t<1$. For a clean chunk $\bz=Q(\ba_{1:H})$,
the conditional probability path is
\begin{equation}
 q_t(\bz_t\mid\bz)
 ={\textstyle \prod_{j=1}^L}\left[
 \alpha(t)\mathbf 1\{z_{t,j}=z_j\}
 +(1-\alpha(t))q_{0,j}(z_{t,j})\right].
 \label{eq:df-path}
\end{equation}
At each position, this path interpolates between the source and the
clean target token. The source $q_0(\cdot)$ may be uniform over $[B]$ or concentrated
on an auxiliary mask token~\citep{sahoo2024simple}, which does not decode to an action bin.
{}
As in DR, the head predicts a clean-token distribution at each position,
but additionally conditions on the current chunk and flow time:
$p_{\theta,j}^{\mathrm{DF}}(\cdot\mid\bz_t,t,o)$.
For each training pair, we sample $t\sim\mathcal U[0,1]$ and
$\bz_t\sim q_t(\cdot\mid\bz)$, and minimize clean-token
cross-entropy~\citep{gat2024discrete}:
\begin{equation}
 \cL_{\mathrm{DF}}
 =\E_{(o,\ba_{1:H})\sim\cD,\,t,\,\bz_t}
 \left[{\textstyle -\frac{1}{L}\sum_{j=1}^L}
 \log p_{\theta,j}^{\mathrm{DF}}(z_j\mid\bz_t,t,o)\right].
 \label{eq:df-prelim}
\end{equation}
At inference, initialize
$\bz_{t_0}\sim q_0$ and use $0=t_0<\cdots<t_N=1$.
At step $n$, draw proposals
$\widetilde z_{n,j}\sim
p_{\theta,j}^{\mathrm{DF}}(\cdot\mid\bz_{t_n},t_n,o)$ and update
each position, writing $\alpha_n=\alpha(t_n)$, as
\begin{equation}
 z_{t_{n+1},j}=
 \begin{cases}
 \widetilde z_{n,j},&\text{with probability }\eta_n,\\
 z_{t_n,j},&\text{otherwise},
 \end{cases}
 \qquad
 \eta_n=\frac{\alpha_{n+1}-\alpha_n}{1-\alpha_n}.
 \label{eq:df-update}
\end{equation}
Finally, the policy returns $\hat \ba_{1:H} \gets (o;\beps_{\mathrm{D}}):=D_Q(\bz_{t_N})$. The source choices and sampler details are in \Cref{app:head-sampler}.
Importantly, while each single DF step predicts independently per position (as in DR), DF introduces position-wise correlation through multi-stage computation of the final output tokens.

\confpar{Classifier-free guidance.}
CFG~\citep{ho2022classifierfree} combines conditional and unconditional
predictions. Let $c$ be the encoding of the observation $o$, and let
$\emptyset$ be a learned null condition of the same shape. Training
replaces $c$ with $\emptyset$ at random, so one head learns both
predictions. We write $\emptyset$ in place of $o$ for a prediction made
with the null condition. For discrete heads, suppressing other inputs,
$p_{\theta,j}^{\mathrm{CFG}}(k\mid o)\propto
p_{\theta,j}(k\mid o)^{1+\lambda}
p_{\theta,j}(k\mid\emptyset)^{-\lambda}$,
normalized over bins, where $\lambda\ge0$ is the guidance weight
($\lambda=0$ means no guidance). We write CFG~$x$ for $\lambda=x$.
DR applies this rule to its single pass and decodes the guided
distributions greedily; DF applies it at every refinement step and
samples clean-token proposals from them. CF applies the same
combination to its velocity,
$(1+\lambda)v_\theta(\bx_t,t,o)-\lambda v_\theta(\bx_t,t,\emptyset)$,
at every integration step. The backbone encodes $o$ once per inference
call, so guidance adds one action-head pass per head evaluation. We do
not guide CR: it outputs a single chunk rather than a distribution, so
guidance has nothing to sharpen and only extrapolates that chunk.
Details are in \Cref{app:head-cfg}.

%% file: body/discrete_vs_continuous.tex
\section{Discrete Action Prediction Can Match Continuous Parametrizations}
\label{sec:discrete-vs-continuous}

In this section, we demonstrate that discrete action prediction matches, and often outperforms continuous action prediction \emph{across} both flow and regression policies. To evaluate performance, and refute possible precision loss under action discretization, we measure (1) task success rate under closed-loop deployment, and (2) mean $\ell_2$ action prediction error on held-out validation data (\Cref{app:env-measures}).

\confpar{Evaluation Protocol.} To demonstrate that our findings are broadly applicable, we consider a wide breadth of tasks; we defer further implementation details to \Cref{app:environments}.
\begin{itemize}[leftmargin=8pt]
    \item \textbf{Single-Task Imitation Learning.} To test whether discrete policies' advantages depend on pretrained VLM, we train separate policies from scratch, without a pretrained VLM backbone, on the Can, Square, Tool Hang and Transport tasks in the RoboMimic suite \citep{robomimic2021}. In particular Tool Hang and Transport are representative of tasks that demand high control precision. Architectures and training settings are in \Cref{app:head-arch-scratch}.
    \item \textbf{VLA Scale Tasks in sim.} We post-train the PaliGemma \cite{beyer2024paligemma} VLM, used by the $\pi$-series models \cite{black2024pi0, physicalintelligence2025pi05}, with new action heads corresponding to each action head parametrization.  As representatives of VLA training, we train on one mixture of eight sources: six RoboCasa365 datasets \citep{robocasa2024, robocasa365}, including MimicGen-generated demonstrations \citep{mandlekar2023mimicgen}, LIBERO-plus \citep{fei2025liberoplus}, and DROID \citep{khazatsky2024droid}. We measure closed-loop success on RoboCasa and LIBERO \citep{liu2023libero}; DROID has no simulator and serves the offline diagnostics. Evaluations on  RoboCasa are grouped into 3 categories: atomic-seen and  composite-seen consisting of atomic and composite tasks, respectively, that are contained in the training data;  and composite-unseen, consisting of tasks that require compositions of in-distribution atomic-seen, but in novel combinations. The training mixture is in \Cref{app:env-vla-data}, the architecture in \Cref{app:head-arch-vla}, and the evaluation protocol in \Cref{app:env-vla}.
    \item \textbf{Real World VLA tasks}. We used the same backbone and action head setting in sim but trained on different real datasets and SFT on three tasks: throw bottles into bins, open pen caps, and turn mug up. The rollout results are provided in \Cref{fig:robot-success}(b)

\end{itemize}

\confpar{Results.} We depict the closed-loop performance of all four policy parametrizations : CR, CF, DR, DF in \Cref{fig:robot-success}(b,c). We see two surprising findings: (i) across  \emph{\textbf{VLA}} and \emph{\textbf{single-task, from-scratch }} training, discrete flow is competitive with, and often slightly surpasses, continuous flow. More surprisingly, (ii) discrete regression \emph{consistently outperforms} continuous regression. \Cref{app:results-backbones} reports results with other backbone settings, and \Cref{app:results-action-space} with other action parameterizations. The fact that discrete \emph{regression} outperforms its continuous counterpart, even on single-task from scratch training, debunks two widely held hypotheses: (a) that precision is a fundamental bottleneck to discrete policies \citep{shafiullah2022behavior}; (b) that  discrete policies are only beneficial in VLA settings, where it has been hypothesized, that discrete predictions are more compatible with token-based VLM backbones \citep{brohan2023rt, driess2026knowledge}.
As further evidence that discrete policies are not fundamentally limited by precision, we present their end-effector reconstruction error in \Cref{fig:jerk-motion}, showing that discrete policies remain competitive (and in some cases slightly outperforms) their continuous counterparts. In fact, in \Cref{fig:jerk-motion}, we show that discrete flow policies in particular enjoy \emph{lower} end-effector error early on in the action chunk generation and along the finer motion regime, suggesting the potential for greater precision when actions are discrete.

\subsection{What Enables the Performance of Discrete Policies?}
\begin{figure}[t]
\centering
\includegraphics[width=0.85\linewidth]{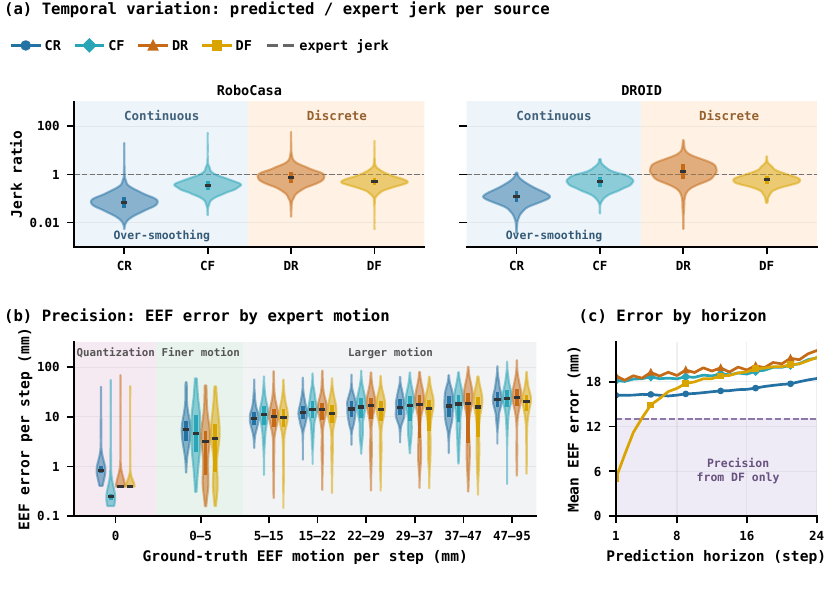}
\caption{
\textbf{Temporal variation and precision on held-out chunks.}
\textbf{(a)} Jerk-ratio violin plots;
Continuous underestimates expert temporal variation.
\textbf{(b)} EEF error, grouped by ground-truth motion.``Quantization'' highlights the discrete heads' static-error floor.
\textbf{(c)} Mean RoboCasa EEF error by horizon; DF has the lowest error in the first steps of the chunk.
More details are in
\Cref{app:responsiveness-heldout}.}
\label{fig:jerk-motion}
\end{figure}
Paradoxically, our previous findings suggest that the opposite of the common wisdom regarding discrete policies might be true: perhaps, discrete policies are \emph{uniformly better} at high-precision action prediction than their continuous counterparts?

However, this belief is not quite correct. \Cref{tab:eef-prediction-error} in \Cref{app:factorization-results} shows that the $\ell_2$ reconstruction errors across all parametrizations are roughly similar, and show \emph{no consistent correlation with task success}. For example, DR exhibits less error than CR on some benchmarks, and more on others, despite \emph{always} enjoying greater closed loop task success.

\confpar{Temporal Variation, via the Jerk Ratio, Reflects Discrete Policy Competitiveness.}
Instead, we propose an alternative explanation for the success of discrete policies:  that they enable \textbf{greater temporal variation} across action chunks than do their continuous counterparts. To make this rigorous,  \Cref{fig:jerk-motion} (a) depicts each chunk's
\textbf{jerk ratio}. \footnote{Recall that \emph{jerk} refers to a the third-time derivative of its actions, and thus reflects the \emph{variability} of actions during a given action chunk.}
For a given action chunk $\ba_{1:H}$, let $\ba_h^{\mathrm{tr}}$
denote the translation coordinates of the end-effector.
We define the  difference approximation to the jerk at step $h$ as $ \Delta^3 \ba_h^{\mathrm{tr}}
 =
 \ba_{h+3}^{\mathrm{tr}}
 -3\ba_{h+2}^{\mathrm{tr}}
 +3\ba_{h+1}^{\mathrm{tr}}
 -\ba_h^{\mathrm{tr}},
 \quad h=1,\ldots,H-3.$
The chunk's mean jerk magnitude and predicted-to-target jerk ratio are
\begin{equation}
 \begin{aligned}
 \mathrm{Jerk}(\ba_{1:H})
 &=
 \frac{1}{H-3}\sum_{h=1}^{H-3}
 \left\|\Delta^3 \ba_h^{\mathrm{tr}}\right\|_2,\quad
 R_{\mathrm{jerk}} =\frac{\mathrm{Jerk}(\hat{\ba}_{1:H})}{\mathrm{Jerk}(\ba_{1:H})}.
 \end{aligned}
 \label{eq:jerk-ratio}
\end{equation}
Thus, jerk ratio of $R_{\mathrm{jerk}} = 1$ means that the learned policy's jerk matches the ground-truth jerk on a training example; $R_{\mathrm{jerk}} < 1$ means jerk is underestimated, so that the policy has \emph{oversmoothed} the training data with respect to time.

\Needspace{20\baselineskip}
\begin{wrapfigure}{r}{0.55\textwidth}
\vspace{-8pt}
\centering
\includegraphics[width=\linewidth]{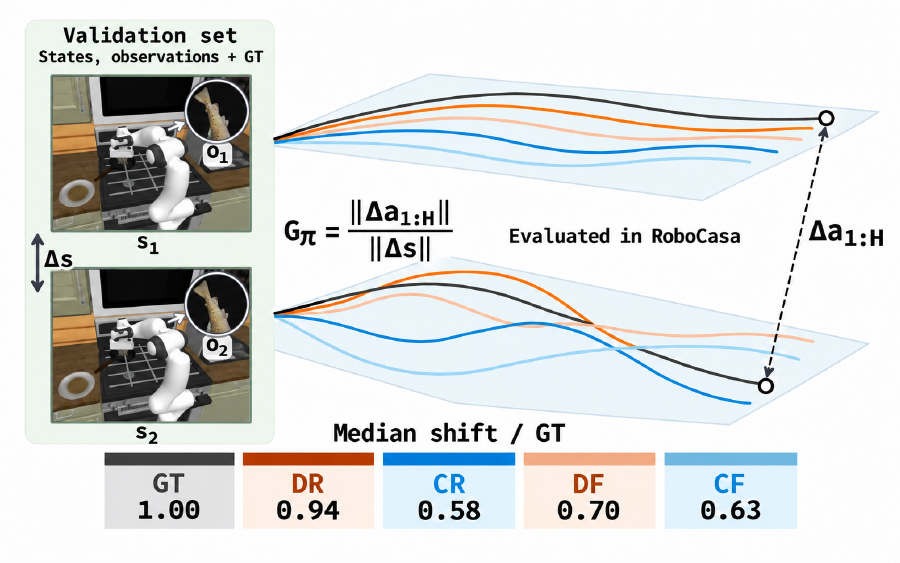}
\caption{\small State responsiveness on held-out RoboCasa
demonstrations: each head's median change in its mean predicted chunk,
divided by the expert's median chunk change, for frame pairs four steps
apart (1.00 matches the expert). Other lags are in
\Cref{tab:expert-response}.}
\label{fig:state-responsiveness}
\end{wrapfigure}
Across all datasets (per-source views in \Cref{fig:variation-by-source}), \Cref{fig:jerk-motion} (a) reveals that continuous regression enjoys a jerk ratio substantially less than $1$, whereas discrete regression is close to (and occasionally above) $1$. This suggests that discrete action modalities can more easily represent large temporal variation in a single pass, likely due to the probabilistic \emph{decoupling} of predictions across time steps. These differences reduce under iterative computation, as \Cref{fig:jerk-motion} (a) shows CF and DF have much closer jerk ratios than do the single-pass counterparts.
To identify differences that persist between DF and CF,
\Cref{fig:jerk-motion} (b,c) reveal that discrete policies better fit other subtle variations: in (b), we see that discrete modalities offer more response to smaller degrees of EEF movement, especially the 0--5\,mm range (note the logarithmic plot scale); in (c) we see that DF in particular fits actions earlier in the chunk more accurately, which require smaller relative motion.
We hypothesize that continuous objectives underweight/over-smooth small action differences compared to categorical prediction.  In \Cref{app:synthetic-structure}, we corroborate these hypotheses in controlled studies of fitting synthetic functions with varying levels of smoothness.

\Needspace{4\baselineskip}
\confpar{DR/DF Exhibit Greater State Responsiveness than CR/CF.}
Beyond temporal variation within a chunk, we examine \emph{state responsiveness}: how much the predicted chunk changes when the observation changes along a real trajectory, formally $\frac{\partial \pi}{\partial o}$, compared with the expert's change (\Cref{app:state-responsiveness}). \Cref{fig:state-responsiveness} shows that both continuous heads produce smaller changes in their mean predicted chunks than their discrete counterparts, underestimating expert change magnitudes on held-out data. DR tracks the expert response most closely, while DF remains closer than CF at every tested state displacement.
On the other hand, additional diagnostics in \Cref{app:robustness} also show that CF stays closer to nearby training actions than DF,\footnote{Closeness to training actions is measured by an expert-calibrated excess distance to the nearest training chunks. This measure, the disturbance protocol, and the sampling baselines are defined in \Cref{app:robustness-protocol}.} corroborating findings from \cite{pan2025noising}. These results suggest that stronger adherence to the training data may come at the cost of responsiveness. DF may offer a better balance, retaining the \emph{disturbance stability} of flow heads, a small change in the predicted distribution after a small simulator disturbance (\Cref{app:disturbance-stability}), while responding more closely to expert changes.

\begin{ssbox}{} \textbf{Takeaway 1:} Discrete policies generally \textbf{match or exceed} the control
performance of their continuous counterparts across the evaluated
testbeds. They obtain stronger performance in both VLA and from-scratch settings. These gains arise from \textbf{better capturing variation} across within-chunk time steps and conditioning observations, rather than outright reconstruction error.
\end{ssbox}

%% file: body/factorization_error.tex
\section{Factorization Error and Iterative Computation in Discrete Policies}
\label{sec:factorization}

Thus far, \Cref{sec:discrete-vs-continuous} establishes that discrete regression (DR) outperforms continuous regression (CR), in part due to better capturing temporal variation, subtle action changes, and response to the input variables.
In this section, we show that the greater temporal reactivity observed in DR comes at the cost of action inconsistency, or what we call \textbf{factorization error} (Finding A), harming policy performance. We then identify that the key interventions which enhance performance---applying  classifier-free-guidance (Finding B) and introducing iterative computation by moving to discrete flow (Finding C) ---correct this inconsistency via complementary mechanisms described below.

\confpar{Factorization Error in Discrete Policies.}
DR predicts a categorical distribution for each action token, giving the
factorized distribution
$p_{\theta}(\bz\mid o)=\prod_j p_{\theta,j}(z_j\mid o)$.
When an observation admits several plausible behaviors, different token
distributions can assign probability to alternatives behaviors independently.
These independent choices can combine incompatible alternatives
across time steps and action dimensions. We refer to this failure to
preserve dependencies between token choices as \textbf{factorization error}.
This may occur even when every token marginal is learned correctly, because
the product of the marginals need not recover the joint distribution of
demonstrated chunks.

\confpar{Evaluation Protocol for Factorization Error.}
To isolate token ambiguity as a potential source of incompatible action tokens,  we vary (i) the number of function evaluations (NFE), that is, the number of refinement steps $N$ in DF, comparing to DR as a baseline, and (ii) the guidance weight $\lambda$ of classifier-free guidance (CFG, \Cref{sec:prelim}). Note that greater NFE provides more opportunity for self-correction, whereas greater CFG ``sharpens'' the generated action distribution.
Due to space constraints, policy performance under varying CFG/NFE is reported in \Cref{app:factorization-nfe-cfg}; in short, performance {improves} with NFE and an appropriate guidance weight.

\begin{figure}[t]
    \centering
    \includegraphics[width=\linewidth]{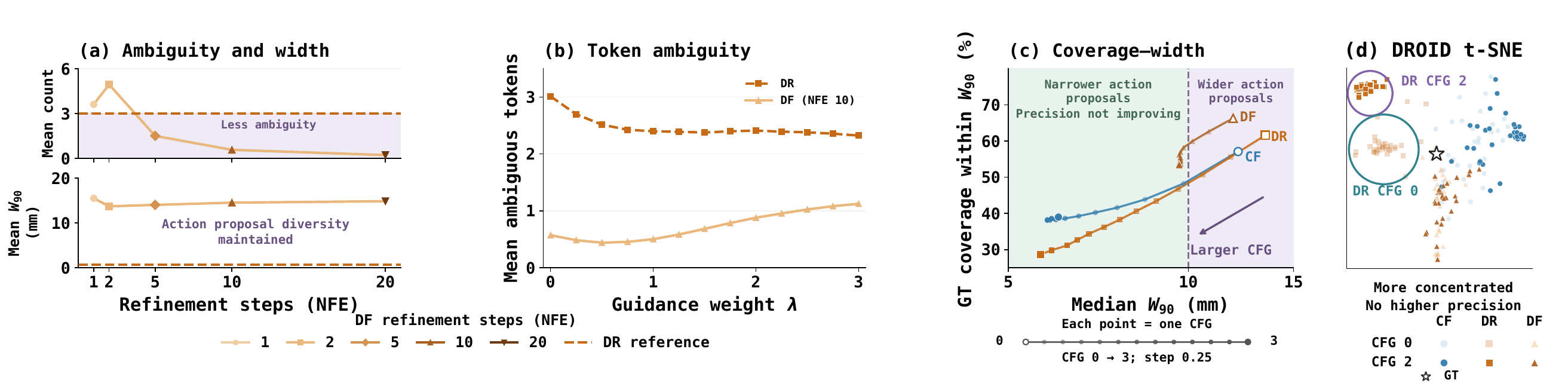}
    \caption{\textbf{Iterative refinement reduces token ambiguity.}
    \textbf{(a,b)} At CFG~0, DF at NFE $1,2,5,10,20$ show
    separated-mode prevalence (a) ambiguity decreasing, action diversity remains
    $\exp(\mathcal{H}(p))$ (b);
    the orange dashed curve is DR.
    \textbf{(c)} Median central-90\% width $W_{90}$ of action proposal and its ground truth coverage, run from CFG~0 to 3.
    \textbf{(d)} DROID t-SNE; the star marks the GT; other sources are in \Cref{fig:cfg-tsne}.
    ; additional diagnostics and
    protocols are in \Cref{app:factorization-method,app:cfg-action-error}.}
    \label{fig:token-ambiguity}
\end{figure}

Here, we focus on  two metrics, deferring precise details to \Cref{app:factorization-method}:
\textbf{1.} \emph{Mode-mixture}: we identify  distributions with at least two
``incompatible'' modes in the same action chunk.
We report the mean count of such modes within one chunk across NFE \Cref{fig:token-ambiguity}(a, top) and across guidance weights \Cref{fig:token-ambiguity}(b). \textbf{2.} \emph{Translation Quantiles}. Next, we measure the \emph{diversity} of actions generated
across repeated policy samples from the same observation. This lets us test (and ultimately rule out) an alternative hypothesis that  DF improves performance over DR by better capturing multimodality. Specifically, we measure the
central-90\% interval width $W_{90}=Q_{0.95}-Q_{0.05}$ for each of the three translation
coordinates per time step, and take the Euclidean norm of the 3 per-axis widths. We plot the norm of median widths against NFE in \Cref{fig:token-ambiguity}(a, bottom) and guidance weight in \Cref{fig:token-ambiguity}(c). Lastly, \Cref{fig:token-ambiguity}(d) qualitatively visualizes action diversity.

\begin{figure}[t]
\centering
\includegraphics[width=\linewidth]{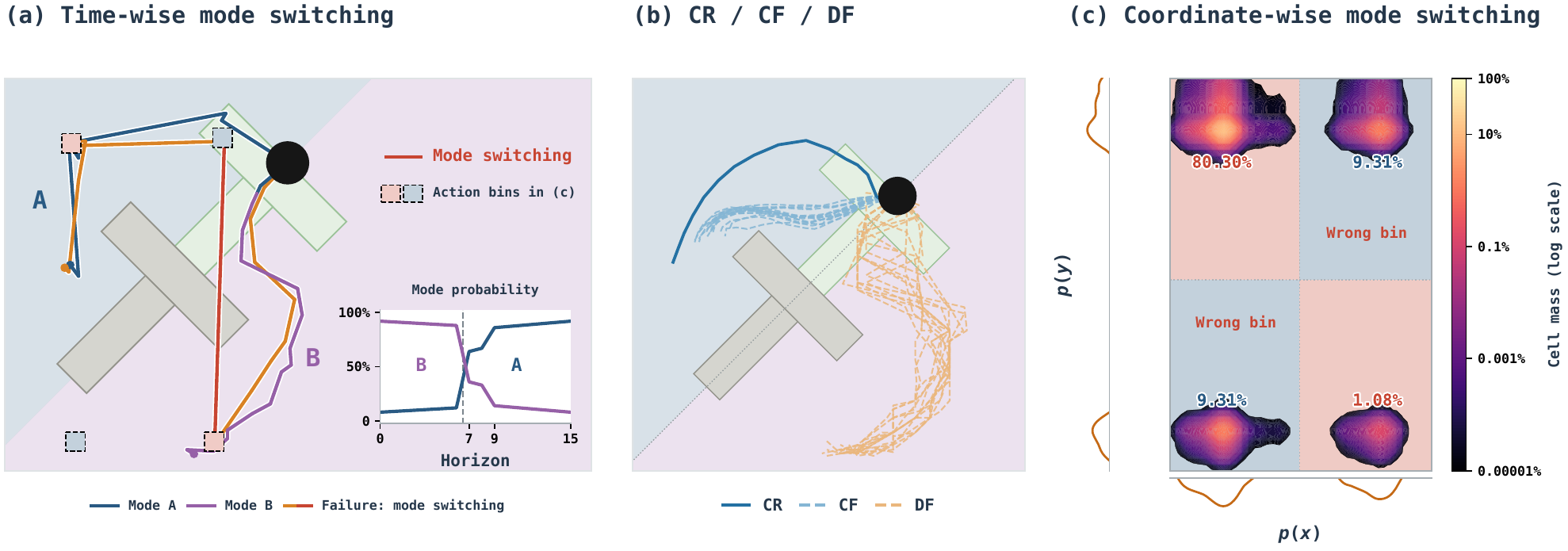}
\caption{\textbf{(a)} Temporal mode switching within an action chunk as relative mode probabilities vary across the prediction horizon;
\textbf{(b)} Action chunks predicted by CR, CF, and DF from the same starting position.
\textbf{(c)} Coordinate-wise factorization within a timestep: independently combining coordinates from two valid modes can produce two additional, incompatible actions.}
\label{fig:pusht-factorization}
\end{figure}

\confpar{Finding A: Factorization Error Across Horizon and Coordinates Can Limit DR.}
In \Cref{fig:token-ambiguity}(a), we show that DR suffers from more mode-mixture
  than DF, even worse at zero CFG. As
  further evidence, ~\Cref{fig:pusht-factorization} illustrates the mechanism in the simple Push-T environment. Here, DR assigns probability to separated movement alternatives, and their relative mass changes across the prediction horizon.
\Cref{app:factorization-robocasa} presents the analogous ambiguity in RoboCasa. These examples confirm the quantitative findings in \Cref{fig:token-ambiguity}, illustrating how factorized decoding
can introduce \emph{spurious} temporal variation, even when individual
  token predictions capture plausible alternatives.

\confpar{Finding B: Classifier-Free Guidance Reduces Ambiguity and Makes DR Competitive with DF.}
As further evidence of the presence of factorization error, we observe that increasing guidance from CFG~0 to CFG~2 raises
pooled RoboCasa success from 27.1\% to 49.2\%
(\Cref{app:factorization-cfg}). This coincides with drastic reduction in the mean-width of the confidence intervals (\Cref{fig:token-ambiguity}(c)), which is consistent with reductions in spurious temporal variation. As noted above, guidance is known to ``sharpen'' the resulting generated distribution \citep{ho2022classifierfree,karras2024guiding}, and thus is likely to be resolving mode ambiguity.
Moreover, \Cref{fig:token-ambiguity}(c,d) shows that CFG reduces coverage of plausible ground-truth EEF positions, suggesting that performance advantages from CFG primarily reflect reduced factorization error rather than improved prediction accuracy. Across sources, CFG almost never improves EEF error by more than 1\,mm (\Cref{app:cfg-action-error}).

\confpar{Finding C: Iterative Computation Aids Discrete Policies via Reducing Factorization Error.}
While CFG guidance is a partial-yet-imperfect remedy of factorization error, we show that iterative computation, as in DF, reduces factorization error further. To support this claim, we find that DF reduces the number of inconsistent-mode predictions per chunk by a factor of 5.28 relative to DR, bringing the average below one at NFE~$\geq 10$ (\Cref{fig:token-ambiguity}(a,top)). This coincides with a 4.7-point increase in RoboCasa success (\Cref{app:factorization-nfe-cfg}). These suggest that iterative computation
makes discrete action predictions more cohesive  than guidance alone. We note that, for continuous action predictions, the benefits of iterative computation are somewhat \emph{opposite}, combating over-smoothing (\Cref{fig:jerk-motion}) and policy under-reactivity (\Cref{fig:pusht-factorization}). The high level role of  iterative computation, however, is the same, in that it performs self-correction on policy actions \citep{pan2025noising}.

\Needspace{9\baselineskip}
\begin{wraptable}{r}{0.5\textwidth}
\vspace{-16pt}
\centering
\small
\setlength{\tabcolsep}{4pt}
\caption{EEF error and RoboCasa success}
\label{tab:autoregression}
\vspace{-6pt}
\begin{tabular}{@{}l>{\bfseries}c>{\color{black!60}}c>{\color{black!60}}c@{}}
\toprule
CFG $0\to2$ & AR & DR (ref.) & DF (ref.) \\
\midrule
EEF (mm) $\downarrow$ & 18.4\,$\to$\,24.4 & 16.4\,$\to$\,16.3 & 14.1\,$\to$\,18.2 \\
Success (\%) $\uparrow$ & 42.8\,$\to$\,48.3 & 27.1\,$\to$\,49.2 & 52.3\,$\to$\,53.9 \\
\bottomrule
\end{tabular}
\vspace{-10pt}
\end{wraptable}
\confpar{Does action diversity explain the performance of DF? (Likely not)}  Despite exhibiting low factorization error, DF exhibits \emph{greater action diversity} than DR, as shown in all \Cref{fig:token-ambiguity} plots.
However, this is unlikely to explain its greater performance. CFG~2 notably reduces DF action diversity, and reduces action diversity in CF and DR considerably more; yet in all three,  performance increases, most notably in DR (see \Cref{app:factorization-nfe-cfg}). Moreover, although performance improves up to NFE$=10$, action diversity remains relatively constant (\Cref{fig:token-ambiguity}(a,bottom) and \Cref{fig:nfe-entropy-diversity}).  These findings suggest that preservation of full action diversity in the policy is not essential for task success.

\confpar{Can temporal autoregression resolve factorization error?} Temporal autoregression provides another way to coordinate action choices across time. We implement a discrete autoregressive (AR) policy that predicts the first action from the observation, then predicts each subsequent action conditioned on the observation and all previously generated actions. Coordinates within each time step are predicted in parallel, requiring 24 sequential predictions for a 24-step chunk. Without CFG, temporal AR achieves 42.8\% success on RoboCasa, exceeding DR's 27.1\% but remaining below DF. However, with CFG, the AR slightly trails both DR and DF (\Cref{tab:autoregression}).

\begin{ssbox}{} \textbf{Takeaway 2:} We identify \textbf{factorization error}, a form of temporal inconsistency, as the cause of DR's relative lack of performance compared to CF and DF. Iterative computation in DF yields maximal performance, but with appropriate CFG, DR's performance is \textbf{surprising close}; this reveals that iterative computation and distribution sharpening are both effective means for combating factorization errors.

\end{ssbox}

%% file: body/geometry_and_tokenization.tex
\section{Additional Structure Provides Limited Gains for Discrete Policies}
\label{sec:geometry-and-tokenization}
\Cref{sec:factorization} links the gains from iterative refinement and guidance to reduced ambiguity in discrete action predictions. In this section, we ask \emph{\textbf{can explicit action geometry or structured tokenization further improve on simple categorical prediction with uniform bins?}} We now consider two mechanisms for adding structure: encoding metric distances between action values during the DF corruption process, or changing how action chunks are tokenized.

\confpar{Explicit Geometry Provides Limited Control Gains.}
We introduce action geometry in two ways. For DR, distance-based label smoothing assigns target probability to nearby action bins as well as the demonstrated bin, with probability decreasing with distance (\Cref{app:head-objectives}). We combine this with logit soft-capping to limit confidence. For DF, we replace the uniform-source mixture path in
\Cref{eq:df-path} with a metric-induced corruption path
\begin{equation}
 q_t^{\mathrm{metric}}(\bz_t\mid\bz)
 =\prod_{j=1}^{L}
 \frac{\exp\!\left[-\beta(\alpha(t))\,d(z_{t,j},z_j)\right]}
 {\sum_{b=1}^{B}\exp\!\left[-\beta(\alpha(t))\,d(b,z_j)\right]},
 \qquad d(k,b)=\frac{|k-b|}{(B-1)/2},
 \label{eq:metric-path-main}
\end{equation}
where $\beta$ is an increasing concentration schedule.
As generation progresses, the metric path concentrates corrupted
tokens near the clean target, whereas the mixture path keeps all
incorrect bins equally likely
(\Cref{app:metric-paths}).
Both DF variants retain the clean-token prediction objective in
\Cref{eq:df-prelim}. Thus, label smoothing changes DR's prediction
targets, whereas the metric path changes DF's corrupted inputs.
The schedules, sampling rates, and results of each variant are provided
in \Cref{app:metric-paths}. The tested DR variant improves RoboCasa success by 2.1 points. The best tested metric-path DF reaches 52.8\% pooled success, compared with 53.9\% for mixture-path DF. Explicit geometry therefore provides a modest gain for DR but does not improve the best observed DF result.

\confpar{Structured Tokenizers Do Not Improve over Naive Binning.}
We next test whether changing the action representation improves control. We evaluate Finite Scalar Quantization (nFSQ)~\citep{mentzer2024finite}, residual vector quantization (RVQ) \citep{zeghidour2021soundstream}, and FAST \citep{pertsch2025fast}, alongside quantile-bin baselines. At their best recorded complete operating points, DF reaches 52.4\% RoboCasa SR with nFSQ, 46.2\% with FAST, and 38.2\% with RVQ (\Cref{app:tokenization-performance}). For DR, quantile bins and nFSQ beat the unguided flat-bin baseline (49.8\% and 41.6\%, against 27.1\%). Against guided flat-bin DR (49.2\%), quantile bins are only marginally better and nFSQ is lower, so the tokenizer matters mainly when DR runs without guidance. However, none of these exceeds our best flat-bin DF result of 53.9\%. The observed gains are limited to DR and do not raise the best recorded performance.

\begin{figure}[t]
\centering
\includegraphics[width=\linewidth]{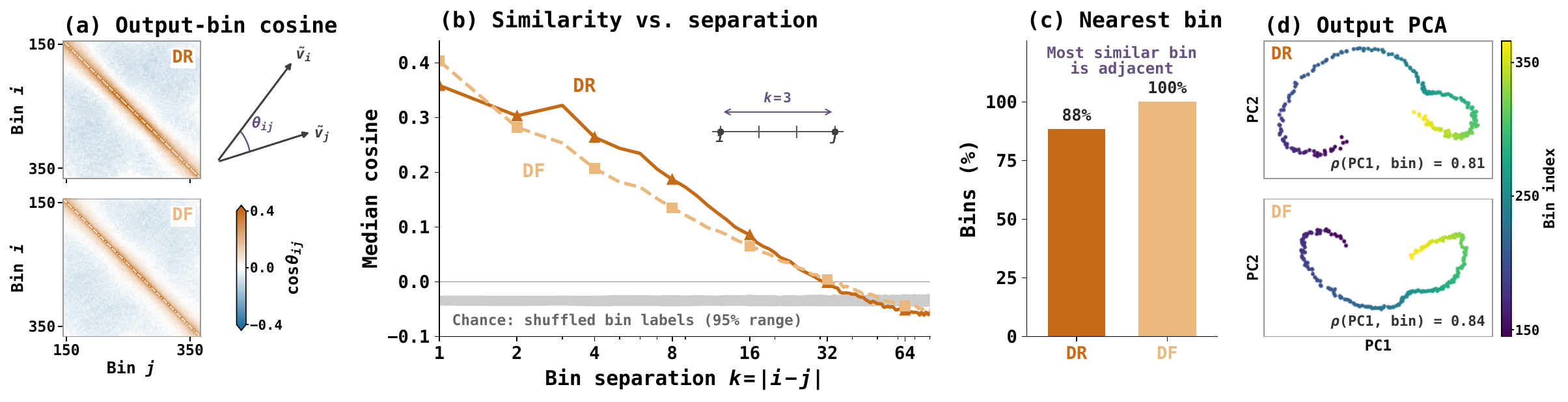}
\caption{\textbf{Discrete heads learn local bin order.} Centered output
embeddings of DR and DF trained on real bimanual robot data, over the 222
bins seen at least 50 times in validation. \textbf{(a)} Cosine similarity
between bins. \textbf{(b)} Median cosine at each bin separation $k$; gray
band: chance under shuffled bin labels. \textbf{(c)} Bins whose most similar
bin is a direct neighbor. \textbf{(d)} First two principal
components, colored by bin index; $\rho$: Spearman correlation of PC1 with
bin index. Details are in \Cref{app:bin-geometry}.}
\label{fig:bin-geometry}
\end{figure}

\confpar{Discrete Policies Can Learn Local Bin Geometry.}
Why does encoding structural information yield little benefit? We hypothesize that discrete policies encode geometry through the \textbf{embedding vectors} associated with the last logit layer. Specifically, each  bin $b$ is associated with a last-layer embedding $\bv_{b}$, shared across all coordinates $j$. For spatially adjacent bins $b,b+1$, we compute the vector cosine $\cos\theta_{b} := \langle \bv_{b},\bv_{b+1}\rangle / ( \|\bv_{b}\|\|\bv_{b+1}\|)$. We find that adjacent bins exhibit high cosine similarities. We study the last-layer embeddings that produce each bin's logit in DR and DF, to see whether the heads only memorize bins as separate labels or learn what each bin represents (\Cref{app:bin-geometry}). Adjacent bins have mean cosine 0.37 for DR and 0.40 for DF, far above a null in which bin labels are randomly permuted (\Cref{fig:bin-geometry}a,b). The similarity decreases with bin separation and falls to chance at about 45 bins, so the learned geometry is local (\Cref{fig:bin-geometry}b). For 88\% (DR) and 100\% (DF) of bins, the most similar other bin is a direct neighbor (\Cref{fig:bin-geometry}c). The first principal component of the embeddings also tracks the bin index (Spearman $\rho=0.81$ for DR and $0.84$ for DF; \Cref{fig:bin-geometry}d). These observations suggest discrete policies do not treat distinct bins as isolated categories, and offer one possible explanation for the limited benefit of imposing that geometry explicitly. Embedding plots, probes, and the selected-checkpoint comparison are provided in \Cref{app:bin-geometry}.

\begin{ssbox}{}{\textbf{Takeaway 3:} Explicit geometry and structured tokenization can improve individual configurations, but neither exceeds the best configuration of DF with the uniform mixture path and flat-bin tokenization. Learned local geometry offers a possible explanation for these limited gains.}
\end{ssbox}

%% file: body/discussion.tex
\section{Discussion and Future Work}
\label{sec:discussion}
We began by asking whether robot actions are fundamentally different from other AI modalities in requiring continuous prediction. Our results challenge this distinction. Although robots operate in a continuous physical world, simple action tokens can support precise control in both simulation and real-world tasks. Their success shows that discrete action prediction is more than a convenient interface to a language model. It is an effective way to learn robot control in its own right.

The key difficulty exposed by our experiments is deciding which token choices belong together. Discrete heads readily express action variation, but independent predictions can combine incompatible behaviors. Iterative refinement and guidance help resolve these conflicts while retaining responsiveness. This connects action prediction to a challenge also studied in diffusion language models: accurate token predictions must be coordinated to form a coherent joint output.

The need for joint consistency also suggests a different objective for action tokenization. Although the tested reconstruction-based tokenizers do not improve on our best uniform-bin DF result, tokenizers designed to make coupled action choices easier to predict may help reduce factorization error. For example, tokens could jointly encode coherent motion across several time steps. Their value may depend on the prediction head and the number of generation steps available. Finding tokenizations suited to discrete regression, iterative flow, or generation with only a few refinement steps is therefore a promising direction. Such tokenizers should be evaluated by the consistency and control performance of the resulting policies, alongside reconstruction quality.

More broadly, these findings support learning action generation within the same token-based framework as perception and reasoning. A natural next step is to test whether sharing representations and prediction machinery across modalities improves co-training and transfer, while preserving the action consistency needed for closed-loop control.These results make a strong case for treating discrete action prediction as a central component of future robot foundation models.

%% file: body/statements.tex
\subsection*{AI Use Statement}
We used generative AI tools to help implement methods, analyze results, make
figures, and edit the paper. We did not use them to form hypotheses or design
the experiments. We have reviewed all AI-assisted work and take responsibility
for the final content of this paper.

\subsection*{Reproducibility Statement}
The evaluation protocol and environments are described in
\Cref{app:environments}, and the action heads, training objectives,
architectures, and training recipes in \Cref{app:action-heads}. Full results
and selected settings are in \Cref{app:results}. We will release code and
model checkpoints upon publication.

%% file: appendix/related_work.tex
\section{Related Work}
\label{app:related}

\paragraph{Action representations.}
Early generalist policies predicted actions as tokens. RT-1 discretizes
each action coordinate into 256 uniform bins \citep{brohan2022rt1},
RT-2 and OpenVLA reuse such bins as tokens of a pretrained
vision-language model \citep{brohan2023rt,Kim2024OpenVLAAO}, and Gato
trains one transformer from scratch on text and on actions in 1024 bins
\citep{reed2022generalist}. The field has since moved to continuous
heads, motivated by the precision of continuous outputs and by the
ability of diffusion models to represent multimodal action
distributions. Diffusion Policy and ACT generate continuous chunks
\citep{chi2023diffusion,zhao2023act}, Octo and RDT-1B use a diffusion
head or backbone \citep{octo2024octo,liu2025rdt}, and most recent VLAs
attach a diffusion or flow-matching module to a pretrained VLM
\citep{li2024cogact,black2024pi0,physicalintelligence2025pi05,nvidia2025gr00t,shukor2025smolvla}
or regress actions directly \citep{kim2025oft}. Methods that keep
discrete actions usually add machinery to address a perceived weakness
of plain bins. Behavior Transformers add a continuous offset to a
discrete cluster to recover precision
\citep{shafiullah2022behavior,cui2023play,lee2024behavior}. FAST,
vector and scalar quantization, and B-spline tokenizers compress chunks
into shorter token sequences
\citep{pertsch2025fast,lee2024behavior,belkhale2024minivla,wang2025vqvla,mete2024quest,zhou2025beast}.
HybridVLA and knowledge insulation combine a discrete token loss with a
continuous action objective in one model
\citep{liu2026hybridvla,driess2026knowledge}. We ask whether this
machinery is needed. Our discrete heads predict plain per-coordinate
uniform bins, and \Cref{sec:geometry-and-tokenization} tests several of these
tokenizers.

\paragraph{Comparisons of discrete and continuous heads.}
The move to continuous heads rests partly on studies that swap the
action head within one model and find continuous heads stronger
\citep{octo2024octo,li2024robovlms,kim2025oft,hou2025dita,wu2026vlanext,ye2026starvlaalpha}.
Octo, RoboVLMs, and OpenVLA-OFT attribute the gap to bin precision. But
these studies change the representation together with other choices.
The discrete head usually decodes tokens autoregressively, and FAST
shows that autoregressive decoding of per-step bins fails at 20\,Hz and
50\,Hz \citep{pertsch2025fast}. The discrete head is often a linear
projection onto the language-model vocabulary while the continuous head
is a new MLP \citep{li2024robovlms,kim2025oft}, and the discrete
baselines in Diffusion Policy and ACT predict one action at a time
\citep{chi2023diffusion,zhao2023act}. When all tokens are decoded in
one pass, the gap shrinks or reverses. RT-1's one-pass bins outperform
a unimodal Gaussian head \citep{brohan2022rt1}, BEAST's parallel tokens
outperform L1 regression on the same B-spline representation
\citep{zhou2025beast}, BAKU's BeT and VQ-BeT heads match an MLP head and
match or beat a diffusion head \citep{haldar2024baku}, and Discrete
Diffusion VLA reports 94.2\% LIBERO success for parallel bins, 81.2\%
for autoregressive bins, and 94.9\% for a continuous diffusion head
\citep{liang2025discrete}. The evidence is still not one-sided:
parallel bins trail L1 regression in OpenVLA-OFT (90.2\% vs.\ 95.3\%)
\citep{kim2025oft}, and the ranking of heads in Dream-VLA changes with
the backbone \citep{ye2026dreamvla}. Separating the representation from
the decoding requires varying them independently.
\citet{pan2025noising} do this for continuous heads, comparing
regression and flow under matched architectures. We extend the
comparison to discrete heads, crossing continuous and discrete
representations with one-pass and iterative generation under one
backbone and training recipe.

\paragraph{Discrete diffusion and flow policies.}
A second line of work keeps discrete actions and replaces
autoregressive decoding with iterative parallel decoding. Discrete
Diffusion VLA decodes 256-bin action tokens inside the VLM by
confidence-ordered unmasking \citep{liang2025discrete}. Some VLAs build
on diffusion language models: LLaDA-VLA and MMaDA-VLA start from one
\citep{wen2025llada,liu2026mmadavla}, and Dream-VLA pretrains one on
discrete action tokens before fine-tuning a flow head
\citep{ye2026dreamvla}. A shared token space also lets one model
generate actions together with text and images. dVLA, UD-VLA, and
MMaDA-VLA denoise these tokens jointly
\citep{wen2025dvla,chen2025udvla,liu2026mmadavla}, and MM-ACT co-trains
the three generation tasks \citep{liang2025mmact}. Later work targets
speed and deployment through block diffusion and caching
\citep{lee2026tbdvla,wang2026blockvla,song2026fastdvla}, structured
masks \citep{yang2026tsmask}, inpainting for asynchronous execution
\citep{wang2026discretertc}, and RL fine-tuning \citep{wu2026dvlarl}.
The approach also extends to policies trained without a VLM
\citep{he2024learning,zhuang2026masked} and to driving
\citep{li2026reflectdrive,xu2025wamdiff,dang2026drivefine}. Closest to
our DF head, DFM-VLA and WAM-Flow use discrete flow matching with a
uniform source and a metric-induced path
\citep{chen2026dfmvla,xu2025wamflow}, and E0 trains a $\pi_0$-style
expert with cross-entropy on Gaussian-noised one-hot bins
\citep{zhan2025e0}. This work focuses on building faster and more
capable models, and leaves open why iterative decoding helps. Several
of these papers find that it beats one-pass decoding in the same model
\citep{liang2025discrete,lee2026tbdvla,liang2025mmact}, but fewer than
half train a continuous head on the same backbone, none of the
manipulation policies applies classifier-free guidance to action
tokens, and DFM-VLA and WAM-Flow do not compare the metric path to a
uniform-source mixture path with all else fixed. \Cref{sec:factorization}
attributes the gain of iterative decoding to reduced factorization
error and adds guidance, and \Cref{sec:geometry-and-tokenization} tests metric-induced
paths.

\paragraph{Discrete diffusion and flow models.}
Our DF head builds on discrete diffusion
\citep{austin2021structured,hoogeboom2021argmax,campbell2022continuous},
in particular its masked variant
\citep{lou2024discrete,sahoo2024simple,shi2024simplified,ou2025absorbing,chang2022maskgit,chang2023muse},
which now scales to large language models
\citep{nie2025large,ye2025dream}, and on discrete flow matching
\citep{gat2024discrete,campbell2024generative}. \citet{shaul2025flow}
extend discrete flow matching to arbitrary paths and propose the
metric-induced path tested in \Cref{sec:geometry-and-tokenization}. Uniform sources and
remasking samplers, which let committed tokens change, are related
alternatives
\citep{schiff2025simple,sahoo2025diffusion,deng2026uniform,wang2025remasking,zhao2025informed}.
Most relevant to us, few-step sampling in these models suffers the same
factorization error as one-pass action heads, because each step
predicts positions independently. ReDi and CoDD reduce this error
\citep{yoo2025redi,li2026factorization}, and \citet{tang2026sampler}
show that it persists even with an exact denoiser. \Cref{sec:factorization}
studies the same error for action chunks.

\paragraph{Discretized control and classification losses.}
\Cref{sec:discrete-vs-continuous} shows that discrete heads can match continuous
ones despite quantization. Work outside imitation learning suggests why
precision is rarely the bottleneck. In reinforcement learning,
per-dimension action bins work well on continuous-control benchmarks
\citep{metz2017discrete,tavakoli2018action,tang2020discretizing,seyde2021bangbang,seyde2023solving,chebotar2023qtransformer},
Dactyl uses only 11 bins per coordinate \citep{openai2020learning}, and
value and reward heads are often trained with cross-entropy over bins
\citep{bellemare2017distributional,hafner2025mastering,hansen2024tdmpc2}.
HL-Gauss, a cross-entropy loss with Gaussian-smoothed bin targets
\citep{imani2018improving}, improves the performance and scaling of
deep RL \citep{farebrother2024stop}. Analyses attribute such gains to
training rather than to modelling the target distribution: easier
optimization in supervised regression \citep{imani2026investigating},
robustness to noisy targets and better representations in RL
\citep{farebrother2024stop}, and different learned features in
two-layer networks \citep{stewart2023regression}. MSE also has a known
failure mode. It averages distinct modes \citep{shafiullah2022behavior}
and smooths over discontinuities \citep{florence2021implicit}, and in
forecasting, models trained with MSE or quantile losses regress toward
the mean or median when the context is uncertain, while cross-entropy
over bins can settle on a mode \citep{yu2026understanding}. For
behavior cloning, the regret with quantized actions and a log-loss
depends on the quantization error only polynomially in the horizon,
given stable dynamics and a probabilistically smooth policy
\citep{cao2026understanding}. Categorical outputs are also standard for
other continuous signals such as pixels and audio
\citep{vandenoord2016pixel,vandenoord2016wavenet}. For VLAs the
evidence is mixed: 512 uniform bins are worse than MSE regression on
CALVIN and better on Meta-World \citep{szot2024grounding}. Our
distance-based label smoothing follows soft ordinal targets
\citep{gao2017deep,diaz2019soft,imani2018improving,chung2026teaching},
and our logit soft-capping follows Gemma~2 \citep{gemma2024gemma2}.

\paragraph{Classifier-free guidance.}
\Cref{sec:factorization} uses classifier-free guidance
\citep{ho2022classifierfree} to sharpen discrete action distributions.
For categorical outputs, guidance reweights each token's distribution
directly. Muse applies it to token logits \citep{chang2023muse}, later
work formalizes it for discrete diffusion and flow
\citep{nisonoff2025unlocking,schiff2025simple,rojas2026improving}, and
on language-model logits it lowers entropy while also reordering the
top tokens \citep{sanchez2024stay}. We apply the Muse rule to each
coordinate's clean-token distribution. In continuous diffusion the
effect is less direct. Guided sampling does not sample the tilted
distribution \citep{chidambaram2024guidance,bradley2025predictor}, and
in simple mixtures large weights push samples toward the edge of the
support \citep{chidambaram2024guidance}. Results on continuous robot
policies are mixed. Guidance helps CogACT \citep{li2024cogact} and
gives a small gain for language-conditioned flow VLAs
\citep{park2026acg}, but it is slightly worse than plain goal
conditioning in BESO \citep{reuss2023goal}, hurts when applied to the
observation \citep{pearce2023imitating}, and makes the RDT-1B arm
unstable \citep{liu2025rdt}. $\pi^{*}_{0.6}$ notes that large weights
push actions to the corners of the support and cause aggressive
motions \citep{physicalintelligence2025pistar}. On discrete action
tokens, the closest prior use is policy contrastive decoding over
OpenVLA's tokens \citep{wu2026policy}. In our experiments, guidance
raises DR success on RoboCasa from 27.1\% to 49.2\% (\Cref{sec:factorization}).

%% file: appendix/environments.tex
\providecommand{\creg}{CR}
\providecommand{\cflow}{CF}
\providecommand{\dreg}{DR}
\providecommand{\dflow}{DF}
\providecommand{\ar}{AR}
\section{Environment Details}
\label{app:environments}
This section gives the evaluation measures, training recipes, and comparison
protocol for the experiments in \Cref{sec:discrete-vs-continuous,sec:factorization,sec:geometry-and-tokenization}.

\subsection{Evaluation Measures and Comparison Protocol}
\label{app:env-measures}
\paragraph{Settings.}
Our evidence spans pretrained
vision--language--action policies, evaluated on RoboCasa and LIBERO and
diagnosed offline on DROID, and policies trained from scratch on RoboMimic
and Push-T. Controlled function fitting uses separate
models with a common trunk architecture to predict deterministic targets from
a scalar input. These settings let us examine precision in robot policies
and study fitting behavior where the target and codec error are known.

\paragraph{Evaluating action quality.}
We distinguish three measurements.
\emph{Prediction error} compares predicted and demonstrated actions, reporting
training fit and held-out error separately. Robot translation errors use
millimeters; Push-T uses pixels.
\emph{Temporal variation}, assessed through jerk, describes variation in the
commanded trajectory.
\emph{Task performance} measures closed-loop success on RoboCasa, LIBERO,
and RoboMimic in Figure~\ref{fig:robot-success}.

Robot evaluations use 5 evaluation seeds with per-head checkpoints and
inference budgets. Synthetic comparisons use 3 initialization seeds.
Formal error definitions, benchmark recipes, and the chronology of the
synthetic studies are in Appendices~\ref{app:environments}
and~\ref{app:synthetic}.

\subsection{From-Scratch Training Experiments}
\label{app:env-scratch}
Single-task RoboMimic policies are state-based and trained from scratch,
without a pretrained vision--language backbone. Their evaluation protocol
and records are in \Cref{app:results-robomimic}.

\subsection{VLA-Scale Experiments}
\label{app:env-vla}
All VLA-scale policies are trained on one shared data mixture
(\Cref{app:env-vla-data}). They share one PaliGemma backbone, action
horizon 24, quantile normalization statistics, and seed~7. Actions are
relative end-effector commands for a Franka arm: target poses with respect to
the end-effector pose at prediction time, with axis-angle rotation.
For each head, we evaluate the last five saved checkpoints, at 110k, 120k,
130k, 140k, and 150k training steps, and report the best of the five. Each
evaluation uses five seeds; we report the mean success rate over the
seeds, and $\pm$ denotes the standard deviation over them. Evaluation replans every
24 steps, and every head runs the same evaluation episodes. The backbone,
optimizer, and batch size are given in \Cref{app:head-arch-vla}.

\paragraph{RoboCasa.}
RoboCasa \citep{robocasa365} is a simulated benchmark of household
manipulation tasks in kitchen scenes. We run 10 trials per task on three
suites: atomic-seen (180 episodes), composite-seen (160), and
composite-unseen (160).

\paragraph{LIBERO.}
LIBERO \citep{liu2023libero} is a simulated benchmark of language-conditioned
tabletop manipulation with a Franka arm. Each suite has 10 tasks with 50
human demonstrations per task. We evaluate closed-loop success on two suites.
In LIBERO-Object, the tasks share one layout and differ in which object the
robot picks and places in a basket. LIBERO-10 (also called LIBERO-Long) has
long-horizon tasks that each chain two subtasks across varied objects,
layouts, and scenes.
The training mixture contains LIBERO-plus demonstrations
(\Cref{app:env-vla-data}), and held-out LIBERO-plus episodes are a
validation source for the offline diagnostics.

\paragraph{DROID.}
DROID \citep{khazatsky2024droid} is a real-world Franka teleoperation dataset
of 76k trajectories collected across 564 scenes and 86 tasks. It is part of
the VLA training mixture. DROID has no simulator, so we do not measure
closed-loop success on it. Held-out DROID chunks instead serve as a validation
source for the offline diagnostics: the jerk ratio and EEF error in
\Cref{fig:jerk-motion}, the t-SNE in \Cref{fig:token-ambiguity}(d), and the
CFG error sweep in \Cref{app:cfg-action-error}.

\subsection{VLA Training Data Mixture}
\label{app:env-vla-data}
The VLA-scale policies train on one mixture of eight sources: six
RoboCasa365 datasets \citep{robocasa365}, LIBERO-plus
\citep{fei2025liberoplus}, and DROID \citep{khazatsky2024droid}
(\Cref{tab:vla-mixture}). All eight share one action space, a 7-dimensional
relative end-effector command for a Franka arm: 3D position and axis-angle
rotation with respect to the end-effector pose at prediction time, and
gripper. One set of quantile normalization
statistics covers the whole mixture.

We filter frames before sampling. For RoboCasa, we drop the frames in which
the robot drives its mobile base and keep the arm-control frames. For DROID,
we keep the successful episodes and remove idle segments. LIBERO-plus is used in full. Each source is sampled with weight
proportional to the square root of its frame count. The RoboCasa MimicGen
source has 116M frames of synthetic demonstrations. To keep it from
dominating the mixture, its frame count is capped at DROID's when we compute
the weight. We hold out 2\% of the episodes of every source, chosen with a
fixed seed. These held-out episodes form the eight validation sources for
the offline diagnostics.

\begin{table}[ht]
\centering\small
\caption{Training mixture for the VLA-scale policies. Frames are counted
after filtering. Weights are proportional to the square root of the frame
count, with the MimicGen count capped at DROID's.}
\label{tab:vla-mixture}
\begin{tabular}{lrr}
\toprule
Source & Frames (M) & Weight\\
\midrule
RoboCasa pretrain atomic & 1.38 & 0.052\\
RoboCasa pretrain composite & 21.89 & 0.206\\
RoboCasa pretrain MimicGen & 116.25 & 0.202\\
RoboCasa target atomic & 2.14 & 0.065\\
RoboCasa target composite-seen & 5.17 & 0.100\\
RoboCasa target composite-unseen & 5.99 & 0.108\\
LIBERO-plus & 2.24 & 0.066\\
DROID & 21.06 & 0.202\\
\bottomrule
\end{tabular}
\end{table}

%% file: appendix/action_heads.tex
\section{Action Head and Objective Details}

\label{app:action-heads}

This appendix gives the training objectives, scalar codec, and sampling
details for the policies introduced in Section~\ref{sec:prelim}.
Table~\ref{tab:heads} summarizes the four head families. Per-experiment
settings are given in Appendices~\ref{app:environments}
and~\ref{app:synthetic}.

\begin{table}[t]
\centering\small
\caption{Four heads: representation crossed with inference computation.
Losses and decoders are described at the head-family level;
per-experiment details are in Appendices~\ref{app:environments}
and~\ref{app:synthetic}.}
\label{tab:heads}
\begin{tabularx}{\linewidth}{ll>{\raggedright\arraybackslash}X>{\raggedright\arraybackslash}X}
\toprule
Head & Output & Training loss & Generation \\
\midrule
CR & real-valued chunk & squared error on the chunk & one pass \\
CF & real-valued chunk & velocity or clean-target MSE
   & $N$ integration steps from noise \\
DR & one token per coordinate
   & token cross-entropy (hard or distance-smoothed)
   & one pass, argmax on every token \\
DF & one token per coordinate
   & clean-token cross-entropy on corrupted chunks
   & $N$ updates, conditioned on the current chunk \\
\bottomrule
\end{tabularx}
\end{table}

\subsection{Continuous Heads}
\label{app:head-continuous}

\paragraph{Continuous regression.}
For $L=HM$, the continuous-regression policy is trained with
\begin{equation}
 \cL_{\mathrm{CR}}
 =\E_{(o,\ba_{1:H})\sim\cD}
 \left[\frac{1}{L}
 \|\pi_\theta^{\mathrm{CR}}(o)-\ba_{1:H}\|_F^2\right],
 \qquad
 \hat\ba_{1:H}=\pi_\theta^{\mathrm{CR}}(o).
 \label{eq:cr-prelim}
\end{equation}
The unrestricted population minimizer is
$\E[\ba_{1:H}\mid o]$. When several actions are plausible under the
same conditioning, this mean need not itself correspond to a demonstrated
action. Auxiliary random inputs are suppressed in this notation; the
expectation also averages over them when present.

\paragraph{Continuous flow.}
Let $\beps\sim\mathcal N(0,\eye)$ be independent Gaussian noise with
the same shape as $\ba_{1:H}$. For $t\sim\mathcal U[0,1]$ and a
continuously differentiable, nondecreasing schedule satisfying
$\alpha(0)=0$ and $\alpha(1)=1$, define the stochastic
interpolant~\citep{albergo2022interpolants}
$I_t=(1-\alpha(t))\beps+\alpha(t)\ba_{1:H}$ and its derivative
$U_t(\ba_{1:H},\beps)=\dot\alpha(t)(\ba_{1:H}-\beps)$.
The velocity field is trained by flow
matching~\citep{lipman2023flow,liu2022rectified}:
\begin{equation}
 \cL_{\mathrm{CF}}
 =\E_{\substack{
 (o,\ba_{1:H})\sim\cD\\
 t\sim\mathcal U[0,1],\;\beps\sim\mathcal N(0,\eye)}}
 \left[\|v_\theta(I_t,t,o)
 -U_t(\ba_{1:H},\beps)\|_F^2\right].
 \label{eq:cf-prelim}
\end{equation}
At inference, use a grid $0=t_0<\cdots<t_N=1$ and Euler updates
\begin{equation}
 \bx_{t_{n+1}}
 =\bx_{t_n}+(t_{n+1}-t_n)v_\theta(\bx_{t_n},t_n,o),
 \qquad \bx_{t_0}=\beps\sim\mathcal N(0,\eye),
 \label{eq:cf-euler-details}
\end{equation}
for $n=0,\ldots,N-1$. The resulting policy is
$\pi_{\theta,N}^{\mathrm{CF}}(o;\beps):=\bx_{t_N}$.

\subsection{Scalar Action Quantization}
\label{app:head-quantization}

Actions are normalized and clipped according to the recipe in
Appendix~\ref{app:environments}. For coordinate $i$, let
$[\ell_i,u_i]$ be its quantization range and
$\Delta_i=(u_i-\ell_i)/B$ its bin width. The ranges may differ across
coordinates. With token position $j=(h-1)M+i$, the uniform quantizer
and bin-center decoder are
\begin{equation}
\begin{aligned}
 Q(\ba_{1:H})_j
 &:=\operatorname{clip}_{[1,B]}
 \left(1+\left\lfloor
 \frac{a_{h,i}-\ell_i}{\Delta_i}\right\rfloor\right),\\
 [D_Q(\bz)]_{h,i}
 &:=\ell_i+(z_j-\tfrac12)\Delta_i.
\end{aligned}
\label{eq:scalar-codec}
\end{equation}
Clipping keeps the index in $[B]$, including at the upper endpoint.
The quantizer flattens the chunk in a fixed order, and the decoder
restores its $H\times M$ shape. Both maps use normalized action
coordinates; normalization is inverted before execution. Thus $B$ is
both the number of bins per coordinate and the action-token vocabulary
size. Unless otherwise stated, $B=512$.
Alternative tokenizations, including quantile-based bins and more
compressive encoders, are described in Appendix~\ref{app:tokenization-performance}.

\subsection{Discrete Training Objectives}
\label{app:head-objectives}

For $\bz=Q(\ba_{1:H})$, discrete regression is trained by tokenwise
cross-entropy:
\begin{equation}
 \cL_{\mathrm{DR}}
 =\E_{(o,\ba_{1:H})\sim\cD}
 \left[-\frac{1}{L}\sum_{j=1}^L
 \log p_{\theta,j}^{\mathrm{DR}}(z_j\mid o)\right].
 \label{eq:dr-prelim}
\end{equation}
The head outputs $L\times B$ logits, with a softmax over bins at each
position~\citep{brohan2022rt1}. Conditional on the model inputs, its
factorized distribution is
$p_\theta^{\mathrm{DR}}(\bz\mid o)=
\prod_{j=1}^L p_{\theta,j}^{\mathrm{DR}}(z_j\mid o)$.
Consequently, maximizing this model's joint probability reduces to
selecting each token's mode~\citep{Devlin2019BERTPO}:
$\hat z_j=\arg\max_{k\in[B]}p_{\theta,j}^{\mathrm{DR}}(k\mid o)$.
The policy returns $\pi_\theta^{\mathrm{DR}}(o):=D_Q(\hat\bz)$.

Discrete flow also predicts the clean token at each position, but
conditions on the corrupted chunk and flow time. We sample
$t\sim\mathcal U[0,1]$ and
$\bz_t\sim q_t(\cdot\mid\bz)$ using
Equation~\eqref{eq:df-path}, and train with clean-token
cross-entropy~\citep{gat2024discrete}:
\begin{equation}
 \cL_{\mathrm{DF}}
 =\E_{\substack{
 (o,\ba_{1:H})\sim\cD\\
 t\sim\mathcal U[0,1]\\
 \bz_t\sim q_t(\cdot\mid\bz)}}
 \left[-\frac{1}{L}\sum_{j=1}^L
 \log p_{\theta,j}^{\mathrm{DF}}(z_j\mid\bz_t,t,o)\right].
 \label{eq:df-details}
\end{equation}
DR variants with distance-smoothed targets are specified below.

\paragraph{Per-benchmark DR loss.}
Four action heads cross representation with decoding: continuous regression
(\creg), continuous flow matching \citep{lipman2023flow} (\cflow), discrete
regression over flat bins with per-token argmax (\dreg), and discrete flow over
flat bins (\dflow). An autoregressive head (\ar) appears in the FAST tokenizer
validity and closed-loop comparisons. The \dreg\ loss differs between benchmarks. On
RoboCasa it is a distance-kernel soft cross-entropy with kernel temperature
$\tau = 0.0025$ in normalized action units, so the neighbor bin's unnormalized
weight is $\exp(-0.0078/0.0025) \approx 0.045$ of the target bin's. On Push-T
it is hard-label cross-entropy. Statements about ``cross-entropy without a
metric'' therefore apply per run.

\subsection{Discrete-Flow Source Distribution and Sampler}
\label{app:head-sampler}

The source factorizes as
$q_0(\bz_0)=\prod_{j=1}^L q_{0,j}(z_{0,j})$.
A uniform source has $q_{0,j}(k)=1/B$ for $k\in[B]$
\citep{gat2024discrete}; a masked source places all mass on an
auxiliary mask token~\citep{sahoo2024simple}. The mask is an input
symbol and does not decode to an action bin. In either case, the
clean-token predictor outputs distributions over $[B]$.

The schedule is continuously differentiable and nondecreasing, with
$\alpha(0)=0$, $\alpha(1)=1$, and $\alpha(t)<1$ for $t<1$.
For the grid $0=t_0<\cdots<t_N=1$, the mixture-path update in
Equation~\eqref{eq:df-update} uses
$\eta_n=(\alpha(t_{n+1})-\alpha(t_n))/(1-\alpha(t_n))$.
Proposals and replacement decisions are independent across positions
conditional on the current chunk. This rule gives an Euler
approximation, in the interpolation coordinate $\alpha$, to the
associated continuous-time Markov chain. Because $\eta_{N-1}=1$,
the final update replaces every position with a clean-token proposal,
so the output contains no mask tokens.

The policy is
$\pi_{\theta,N}^{\mathrm{DF}}(o;\xi):=D_Q(\bz_{t_N})$, where
$\xi$ collects randomness in initialization, proposals, and replacement
decisions. We suppress $N$ and $\xi$ when clear from context.
For a uniform source, tokens updated at earlier steps may be revised
later. Subsequent predictions condition on the updated chunk, allowing
the generation process to capture dependencies across positions despite
the conditional independence of updates within each step.

\subsection{Classifier-Free Guidance}
\label{app:head-cfg}

CFG~\citep{ho2022classifierfree} combines a prediction conditioned on the
observation $o$ with one made from a null condition $\emptyset$.

\paragraph{Condition and null condition.}
Let $c$ be the encoding of the observation $o$. For the VLA policies,
$c$ is the set of 64 condition tokens that the resampler produces from
the backbone states and the proprioception token
(\Cref{app:head-arch-vla}), so it carries the images, the instruction,
and the proprioceptive state. The null condition $\emptyset$ is a
learned set of 64 tokens with the same shape. During training, each
example's condition tokens are replaced by $\emptyset$ with probability
0.1, so the same head learns the conditional prediction and an
unconditional prediction that sees no observation. All other head inputs
(the current chunk, the flow time, and the auxiliary random inputs) are
unchanged. We write $\emptyset$ in place of $o$ for predictions made with the
null condition, as in $p_{\theta,j}(k\mid\emptyset)$,
$v_\theta(\bx_t,t,\emptyset)$, and $\pi_\theta^{\mathrm{CR}}(\emptyset)$.

\paragraph{Discrete heads.}
Suppressing other model inputs, discrete heads use
\begin{equation}
 p_{\theta,j}^{\mathrm{CFG}}(k\mid o)
 \propto p_{\theta,j}(k\mid o)^{1+\lambda}
 p_{\theta,j}(k\mid\emptyset)^{-\lambda},
 \qquad k\in[B],\quad\lambda\ge0,
 \label{eq:cfg-head-details}
\end{equation}
with normalization over bins. Equivalently, if
$s_{\theta,j}(o)\in\R^B$ and $s_{\theta,j}(\emptyset)\in\R^B$ are the
conditional and unconditional logits at position $j$, apply softmax to
$(1+\lambda)s_{\theta,j}(o)-\lambda s_{\theta,j}(\emptyset)$.
The rule multiplies $p_{\theta,j}(k\mid o)$ by
$\bigl(p_{\theta,j}(k\mid o)/p_{\theta,j}(k\mid\emptyset)\bigr)^{\lambda}$,
so it moves mass toward bins that the observation makes more likely than
the null condition does. Guidance is usually paired with an iterative
sampler. Because DR is trained with the same condition dropout, its
single forward pass also gives unconditional token distributions, so we
apply the rule directly to DR's one-pass logits. Both DR passes receive
the same auxiliary random input tokens, and DR takes the argmax of each
guided distribution. DF applies the rule at every refinement step: at
step $n$, both passes read the same current chunk $\bz_{t_n}$ and time
$t_n$, and DF samples clean-token proposals from the guided
distributions before the update in Equation~\eqref{eq:df-update}.

\Needspace{6\baselineskip}
\paragraph{Continuous flow.}
At each Euler step, CF evaluates the head at the same state
$\bx_{t_n}$ and time $t_n$, once with $o$ and once with $\emptyset$, and the
Euler update of Equation~\eqref{eq:cf-euler-details} uses the guided
velocity in place of $v_\theta(\bx_{t_n},t_n,o)$. The guided velocity is
\begin{equation}
 v_\theta^{\mathrm{CFG}}(\bx_{t_n},t_n,o)
 =(1+\lambda)v_\theta(\bx_{t_n},t_n,o)
 -\lambda v_\theta(\bx_{t_n},t_n,\emptyset).
 \label{eq:cfg-velocity}
\end{equation}

\paragraph{Why CR is not guided.}
Guidance reweights a distribution. For DR and DF it tilts each token's
distribution, and for CF it changes the velocity field that carries
noise to actions. CR has no distribution to reweight: it outputs one
chunk, its estimate of $\E[\ba_{1:H}\mid o]$, and its unconditional
output estimates the dataset mean $\E[\ba_{1:H}]$. The same combination,
$(1+\lambda)\pi_\theta^{\mathrm{CR}}(o)-\lambda\pi_\theta^{\mathrm{CR}}(\emptyset)$,
only moves the predicted chunk along
$\E[\ba_{1:H}\mid o]-\E[\ba_{1:H}]$, further from the dataset mean.
When several behaviors are plausible, the conditional mean already
averages them, and moving it does not select one of them. We therefore
evaluate CR without guidance.

\paragraph{Inference cost.}
The backbone and resampler encode $o$ once per inference call. Because
$\emptyset$ is a fixed learned tensor, the unconditional prediction
needs no backbone pass, only a second pass of the action head. We report
NFE as the number of steps $N$ for DF and CF. Guidance doubles the
action-head passes, from one to two for DR and from $N$ to $2N$ for DF
and CF, and leaves the backbone cost unchanged. Under this convention,
$\lambda=0$ gives the unguided conditional prediction.

\subsection{Omitted Details for Policy Architecture}
\label{app:head-arch}
The heads use the same backbone design. CR and DR receive auxiliary
random inputs independent of the target action chunk, so random inputs
are present for both regression and flow heads. These auxiliary inputs
are suppressed in the preceding notation
Within each setting, all heads see the same observations and training data.
This section gives the networks that map these observations to actions; the
losses and samplers are given in \Cref{app:head-continuous,app:head-objectives}.

\subsubsection{Single-Task From-Scratch Architectures}
\label{app:head-arch-scratch}
RoboMimic policies observe the object state and the end-effector position,
orientation, and gripper state of each arm, over the last two time steps.
An MLP encodes the flattened observation. Each policy predicts $H=10$
actions and executes 8 before replanning. An action is a 10-dimensional
relative end-effector command: 3D position, 6D rotation, and gripper
(20 dimensions for the two-arm Transport task). Actions are normalized per
dimension to $[-1,1]$. Discrete heads use $B=512$ bins per coordinate.
\Cref{tab:arch-scratch} lists the two network families.

\begin{table}[!htb]
\centering\small
\caption{Networks for the single-task RoboMimic policies. Both families are
DiT-style transformers conditioned through adaptive layer norm (adaLN) on
the observation embedding and the flow time. Parameter counts include the
observation encoder and are given for single-arm tasks.}
\label{tab:arch-scratch}
\begin{tabular}{lll}
\toprule
& CR, CF & DR, DF\\
\midrule
Blocks & 8 encoder + 8 decoder & 12\\
Width / heads / MLP width & 256 / 4 / 1024 & 256 / 4 / 1024\\
Action tokens & one per time step (10) & one per time step and coordinate (100)\\
Action input & MLP on the 10-d action & learned embedding of $B$ bins\\
Action output & linear to 10 values & logits over $B$ bins\\
Observation MLP width & 256 & 512\\
Dropout & 0 & 0.1\\
Parameters & 16.4M & 15.8M\\
\bottomrule
\end{tabular}
\end{table}

CF integrates its velocity field with 9 Euler steps. DF runs 100 update
steps with the mixture path $\kappa_t=t^3$, followed by one argmax
prediction. All models are trained for 300k steps with AdamW (learning rate
$10^{-4}$, weight decay $10^{-5}$, gradient clipping at 10) and an
exponential moving average of the weights with decay 0.995, which is used
for evaluation. Continuous heads use batch size 1024 and cosine decay;
discrete heads use batch size 512, a 5\% linear warmup, and cosine decay.
Each run uses one L40S GPU. The synthetic-fitting trunk is described in
\Cref{app:synthetic}.

\begin{table}[tp]
\centering\small
\caption{Action heads and training settings for the VLA-scale policies.
Parameter counts cover the action head and resampler; the backbone adds
2.92B parameters.}
\label{tab:arch-vla}
\begin{tabularx}{\linewidth}{l*{4}{>{\raggedright\arraybackslash}X}}
\toprule
& CR & CF & DR & DF\\
\midrule
\multicolumn{5}{l}{\textit{Shared by all four heads}}\\
Backbone & \multicolumn{4}{l}{PaliGemma-3B from the $\pi_0$ weights, fully fine-tuned}\\
Condition tokens & \multicolumn{4}{l}{64, from a 2-block resampler (width 1536)}\\
Blocks / width / heads / MLP & \multicolumn{4}{l}{8 / 1536 / 12 / 6144}\\
Action chunk & \multicolumn{4}{l}{$H=24$ steps, $M=7$ coordinates, one sequence element per coordinate}\\
\midrule
\multicolumn{5}{l}{\textit{Shared within each representation}}\\
Action representation
  & \multicolumn{2}{>{\raggedright\arraybackslash}p{0.32\linewidth}}{real-valued vector $\ba_{1:H}\in\R^{H\times M}$}
  & \multicolumn{2}{>{\raggedright\arraybackslash}p{0.32\linewidth}}{token sequence $\bz=Q(\ba_{1:H})\in[B]^{HM}$, $B=512$}\\
Action input
  & \multicolumn{2}{>{\raggedright\arraybackslash}p{0.32\linewidth}}{real-valued vector of the same shape}
  & \multicolumn{2}{>{\raggedright\arraybackslash}p{0.32\linewidth}}{token sequence of the same length}\\
Action output
  & \multicolumn{2}{>{\raggedright\arraybackslash}p{0.32\linewidth}}{one real value per element}
  & \multicolumn{2}{>{\raggedright\arraybackslash}p{0.32\linewidth}}{logits over $B$ bins per element}\\
\midrule
\multicolumn{5}{l}{\textit{Per head}}\\
Input content & random input & interpolant $I_t$ & random input
  & corrupted tokens $\bz_t$, uniform source\\
Output target & clean chunk & velocity & clean tokens & clean tokens\\
Training loss & squared error & flow matching
  & distance-smoothed cross-entropy & clean-token cross-entropy\\
Generation & one pass & Euler integration & one pass, argmax
  & $N$ updates, $\alpha(t)=t^3$\\
CFG & no & yes & yes & yes\\
Parameters & 367M & 367M & 369M & 371M\\
\midrule
\multicolumn{5}{l}{\textit{Training, shared by all four heads}}\\
Training steps & \multicolumn{4}{l}{150k}\\
Global batch size & \multicolumn{4}{l}{512 (64 per GPU on 8 B200 GPUs)}\\
Optimizer & \multicolumn{4}{l}{AdamW, learning rate $7.5\times10^{-5}$, $\beta=(0.9,0.95)$, weight decay $10^{-4}$}\\
Gradient clipping & \multicolumn{4}{l}{1}\\
Learning-rate schedule & \multicolumn{4}{l}{linear warmup, then cosine decay to $0.1\times$ the peak}\\
Precision & \multicolumn{4}{l}{bf16}\\
\bottomrule
\end{tabularx}
\end{table}

\subsubsection{VLA-Scale Architectures}
\label{app:head-arch-vla}
All four heads share a PaliGemma-3B backbone \citep{beyer2024paligemma}:
a SigLIP So400m/14 image encoder and a Gemma-2B language model. We
initialize it from the $\pi_0$ vision--language weights
\citep{black2024pi0} and fine-tune it fully. The backbone reads up to three
$224\times224$ camera images (256 tokens each) and up to 48 text tokens with
bidirectional attention. A resampler with 64 learned queries (2 blocks,
width 1536) compresses the final-layer backbone states, together with one
proprioception token, into 64 condition tokens.

The action head is a separate transformer with 8 blocks, width 1536,
12 attention heads, and MLP width 6144. Each block applies self-attention
over the action sequence, cross-attention to the 64 condition tokens, and an
MLP. The head does not share attention layers with the backbone. It
processes the $H=24$ action chunk as a sequence with one element per action
coordinate ($M=7$: 3D position change, axis-angle rotation change, and
gripper). Continuous heads represent each element as a real value; discrete
heads represent it as a token over $B$ bins. Learned position and coordinate
embeddings and Fourier features of the flow time are added to every element.
\Cref{tab:arch-vla} lists what all heads share, what each representation
shares, what differs per head, and the training settings.

Bins are uniform over the normalized action range. DF uses a uniform
source, and the reported DF policy uses the schedule $\alpha(t)=t^3$.
The condition is dropped with probability 0.1 and replaced by a learned
null condition, which enables CFG at inference (\Cref{app:head-cfg}).
Training settings are listed in \Cref{tab:arch-vla}, and the evaluation
protocol, including checkpoint selection and seeds, is given in
\Cref{app:env-vla}.

%% file: appendix/results.tex
\section{Full Performance Results}\label{app:results}

\subsection{RoboMimic Results}
\label{app:results-robomimic}
RoboMimic policies are state-based, trained from scratch on each task,
and use relative actions. For each task and head family,
\Cref{fig:robot-success} reports the best success rate over tokenizer
(flat or quantile bins), flow variant, and guidance weight. Each
evaluation uses 32 episodes.

\begin{figure}[ht]
\centering
\includegraphics[width=0.85\linewidth]{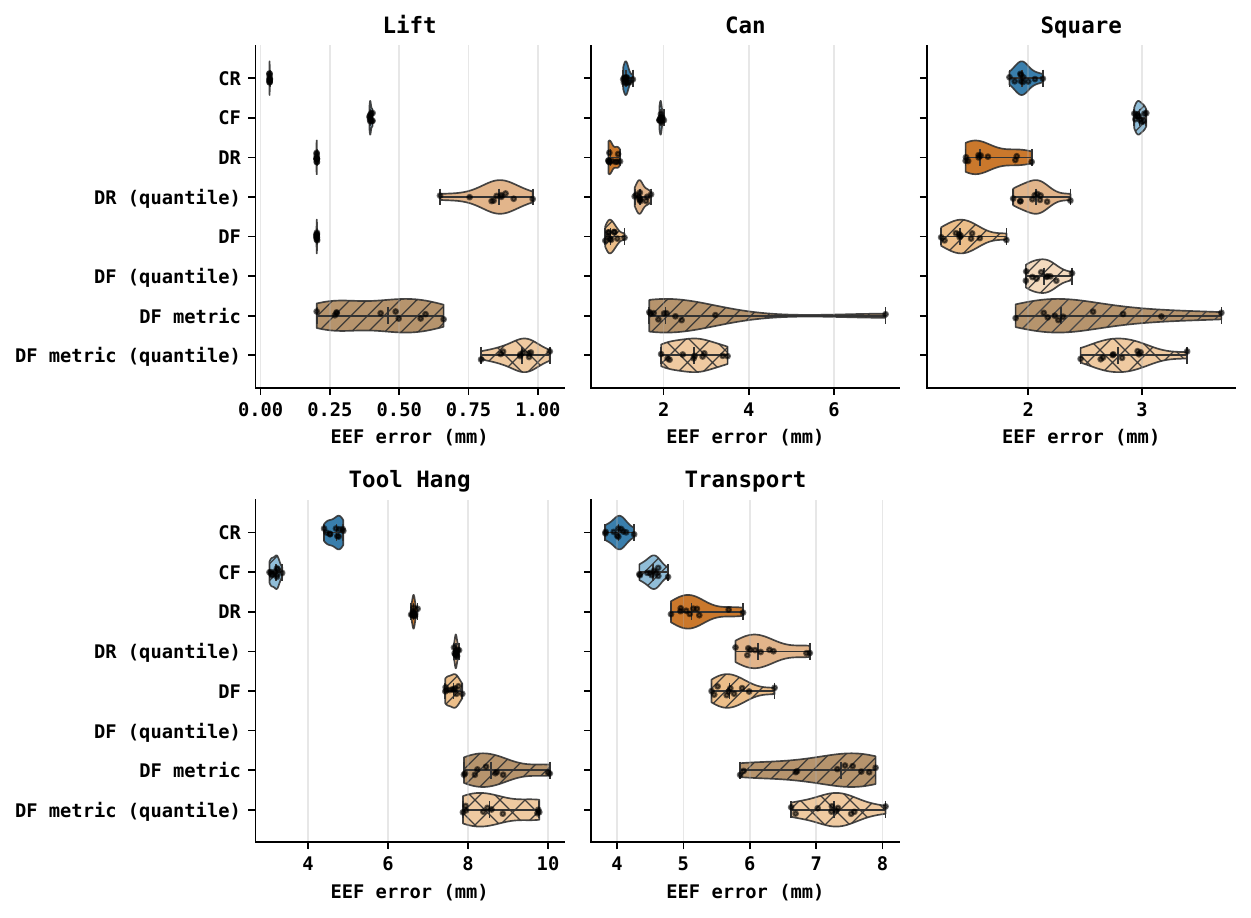}
\caption{\textbf{EEF position error on RoboMimic training samples.} Each
violin shows one run's RMS EEF position error (mm) over its last 10
evaluations; dots are individual evaluations and the vertical line marks
the median. The spread reflects late-training variation of the metric. DF denotes the mixture path and DF metric the
metric-induced path (\Cref{app:metric-paths}); ``quantile'' marks
equal-frequency bins, otherwise bins are uniform.}
\label{fig:robomimic-eef-error}
\end{figure}

\Cref{fig:robomimic-eef-error} reports how closely each run fits its training
data. For each run, we show the end-effector (EEF)
position error on training samples at its last 10 evaluations.

\subsection{Action Parameterization and Action Space}
\label{app:results-action-space}
\begin{figure}[!t]
\centering
\includegraphics[width=\linewidth]{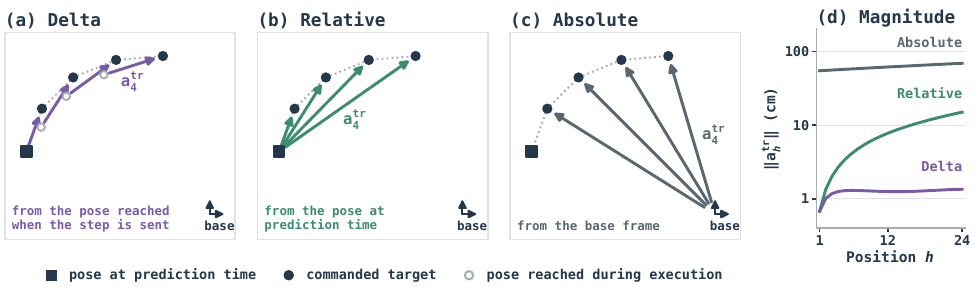}
\caption{\textbf{Delta, relative, and absolute actions.} (a)--(c) One
planned chunk in the robot base frame. Each arrow is the translation
$\ba^{\mathrm{tr}}_h$ that the parameterization encodes for one command,
drawn from its reference pose in \Cref{eq:action-parameterization}.
Rotations use the same reference poses. (d) Translation magnitude encoded
along a 24-step chunk of an illustrative motion, on a logarithmic axis. The
trajectories are illustrative, and (a)--(c) are not to scale.}
\label{fig:action-parameterization}
\end{figure}

\textbf{Both the VLA and the single-task RoboMimic experiments in the main
paper use relative actions in end-effector space.} This section defines
relative actions together with two alternatives, delta and absolute
actions, and compares all three for the four VLA heads on RoboCasa
(\Cref{fig:action-parameterization,tab:ablation-action-space}). It then
compares relative actions in joint space and end-effector space
(\Cref{tab:ablation-joint-eef}).

\paragraph{Definitions.} Let the
chunk $\ba_{1:H}$ be predicted at environment step $t$, and let command $h$
be sent at step $t+h-1$ with commanded target pose $(\bp^\star_h,R^\star_h)$,
where $\bp^\star_h\in\R^3$ and $R^\star_h\in\mathrm{SO}(3)$ are expressed in
the robot base frame. Let $(\bp_k,R_k)$ be the end-effector pose reached at
step $k$. Each parameterization encodes the target with respect to a
reference pose $(\bar\bp_h,\bar R_h)$:
\begin{equation}
\ba^{\mathrm{tr}}_h=\bp^\star_h-\bar\bp_h,\qquad
\ba^{\mathrm{rot}}_h=\operatorname{Log}\bigl(R^\star_h\bar R_h^{\top}\bigr),\qquad
(\bar\bp_h,\bar R_h)=
\begin{cases}
(\bp_{t+h-1},R_{t+h-1}) & \text{delta},\\
(\bp_t,R_t) & \text{relative},\\
(\bzero,\eye) & \text{absolute}.
\end{cases}
\label{eq:action-parameterization}
\end{equation}

\Needspace{12\baselineskip}
\begin{wraptable}{r}{0.5\linewidth}
\vspace{-12pt}
\centering\small
\setlength{\tabcolsep}{3pt}
\caption{Action-space comparison on RoboCasa. Success rate (\%) of the
best of the last five checkpoints, as mean $\pm$ standard deviation over
five seeds. Bold marks the highest success rate in each row.}
\label{tab:ablation-action-space}
\vspace{-6pt}
\begin{tabular}{@{}lccc@{}}
\toprule
Head & Delta & Relative & Absolute \\
\midrule
DF & $51.6 \pm 2.6$ & $\mathbf{53.9 \pm 1.2}$ & $47.0 \pm 1.3$ \\
CF & $50.2 \pm 3.5$ & $\mathbf{52.7 \pm 3.4}$ & $39.0 \pm 2.1$ \\
DR & $\mathbf{50.2 \pm 2.9}$ & $49.2 \pm 1.4$ & $46.8 \pm 1.9$ \\
CR & $37.3 \pm 1.8$ & $\mathbf{38.8 \pm 3.3}$ & $17.4 \pm 3.1$ \\
\bottomrule
\end{tabular}
\vspace{-10pt}
\end{wraptable}
\noindent Here $\operatorname{Log}:\mathrm{SO}(3)\to\R^3$ returns the axis-angle
vector, so $R^\star_h=\operatorname{Exp}(\ba^{\mathrm{rot}}_h)\,\bar R_h$.
The gripper coordinate $\ba^{\mathrm{g}}_h$ is the same in all three.
Delta actions are per-step pose increments, measured from the pose the robot
has reached when the command is sent. Relative actions measure every target
in the chunk from the end-effector pose at prediction time; all policies
in the main paper use them. Absolute actions are target poses in
the robot base frame.

Three properties follow from \Cref{eq:action-parameterization}. At $h=1$,
delta and relative actions coincide. Relative and absolute actions fix every
target of the chunk at prediction time, while a delta target also depends on
the pose reached during execution. Relative actions start near zero and grow
along the chunk, while absolute actions also carry the robot's position in
the workspace, so their range is set by the workspace rather than by the
motion (\Cref{fig:action-parameterization}(d)).

\paragraph{Results.} Each entry of \Cref{tab:ablation-action-space} uses
the decoding setting (CFG and, for the flow heads, NFE) chosen separately
for that head and action space. Relative actions give the highest success
for DF, CF, and CR. DR does slightly better with delta actions (50.2\%
against 49.2\%). Absolute actions give the lowest success for every head.
The drop from relative to absolute actions is smaller for the discrete
heads, 2.4 points for DR and 6.9 for DF, than for the continuous heads,
13.7 points for CF and 21.4 for CR.
\Needspace{11\baselineskip}
\begin{wraptable}{r}{0.46\linewidth}
\vspace{-12pt}
\centering\small
\setlength{\tabcolsep}{3pt}
\caption{Relative actions in joint space and end-effector space on
RoboCasa. Success rate (\%), mean $\pm$ standard deviation over five
evaluation seeds.}
\label{tab:ablation-joint-eef}
\vspace{-6pt}
\begin{tabular}{@{}lcc@{}}
\toprule
Head & Joint space & EEF space\\
\midrule
DF & $54.3 \pm 1.5$ & $53.9 \pm 1.2$\\
CF & $51.9 \pm 2.8$ & $52.7 \pm 3.4$\\
DR & $49.8 \pm 1.9$ & $49.2 \pm 1.4$\\
CR & $38.1 \pm 2.6$ & $38.8 \pm 3.3$\\
\bottomrule
\end{tabular}
\vspace{-10pt}
\end{wraptable}
\paragraph{Joint-space versus end-effector-space actions.}
We compare the four heads using relative actions in joint space and in
end-effector space. Relative joint actions encode each target joint
configuration as an offset from the configuration at prediction time,
$\ba_h^{\mathrm{joint}}=\bq_h^\star-\bq_t$. Relative end-effector actions
encode target translations and rotations with respect to the
end-effector pose at the same prediction time, following
\Cref{eq:action-parameterization}. Both representations therefore use a
fixed reference throughout the predicted chunk. The two action spaces
give nearly the same success for every head; the largest difference is
0.8 points, for CF, which is within one standard deviation
(\Cref{tab:ablation-joint-eef}). In both, DF and CF achieve the highest
success rates, DR remains competitive, and CR performs less well. The
benefits of flow-based prediction, together with the strong performance
of discrete regression, therefore extend across joint-space and
end-effector-space control.

\Needspace{14\baselineskip}
\subsection{Backbone Variants}
\label{app:results-backbones}
\begin{wraptable}[9]{r}{0.46\linewidth}
\vspace{-14pt}
\centering\small
\setlength{\tabcolsep}{3pt}
\caption{RoboCasa success rate (\%) with other backbones, mean $\pm$
standard deviation over five seeds.}
\label{tab:backbone-results}
\vspace{-6pt}
\begin{tabular}{@{}lccc@{}}
\toprule
Head & CFG / NFE & \begin{tabular}[b]{@{}c@{}}$\pi_{0.5}$\\fine-tuned\end{tabular} & \begin{tabular}[b]{@{}c@{}}$\pi_0$\\frozen\end{tabular}\\
\midrule
CR & 0 / -- & $42.4 \pm 1.2$ & $10.2 \pm 0.6$\\
CF & 2 / 20 & $58.2 \pm 2.4$ & $13.4 \pm 0.6$\\
DR & 2 / -- & $55.8 \pm 3.1$ & $10.2 \pm 1.0$\\
DF & 2 / 5 & $60.2 \pm 1.9$ & $11.2 \pm 0.8$\\
\bottomrule
\end{tabular}
\vspace{-10pt}
\end{wraptable}
We also train each head with two other backbone settings: a fine-tuned
backbone initialized from the $\pi_{0.5}$ weights, and a frozen backbone
initialized from the $\pi_0$ weights. Only the action head and resampler
are trained in the frozen setting. All other settings follow
\Cref{app:head-arch-vla}. \Cref{tab:backbone-results} reports, for each head,
the operating point with the highest pooled success over checkpoints,
guidance weights $\lambda\in\{0,2\}$, and NFE $\in\{5,10,20\}$ for the flow
heads.

%% file: appendix/responsiveness.tex
\section{Temporal Variation and Policy Responsiveness Experiments}
\label{app:responsiveness}
This section supplements \Cref{sec:discrete-vs-continuous}.

\subsection{Temporal Variation on Held-Out Chunks}
\label{app:responsiveness-heldout}
\paragraph{Figure~\ref{fig:jerk-motion}: offline replay.}
Each of eight sources supplies 1,000 matched 24-step chunks, giving 32,000
model--chunk observations. All heads use CFG~0; both flow heads use NFE~10.
Jerk is the mean norm of third finite differences of translation commands,
without physical-time normalization. The ratio divides predicted jerk by
target jerk after the same normalization and clipping.

The logarithmic scatter and its coordinate-wise medians omit 488 exact-zero
ratios and 400 zero-target-denominator observations. All positive finite
ratios remain. EEF violins retain all observations, with independently
normalized density widths, medians, and interquartile ranges. Their horizontal
positions are for layout only. Vertical scales vary by source.
Including zero numerators wherever the denominator is positive, CR's
per-source median ratio ranges from 0.036 to 0.286, and CF's from 0.231 to
0.549. CR lies below one on 98.8--99.8\% of defined-ratio chunks and CF
on 85.0--98.1\%. The corresponding median ranges are 0.294--1.313 for DR
and 0.343--0.589 for DF. DF's EEF-error 95th percentile is below DR's on
seven sources, with LIBERO-plus the exception.

\begin{figure}[ht]
\centering
\includegraphics[width=\linewidth]{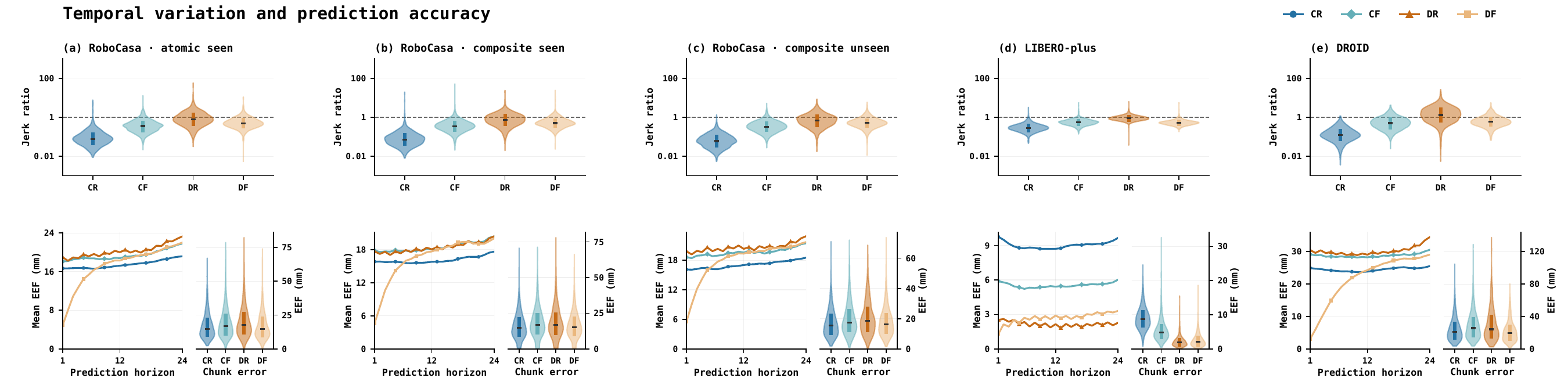}
\caption{\textbf{Per-source view of Figure~\ref{fig:jerk-motion}.}
Top: jerk-ratio violins for each of the five displayed sources. Bottom: mean EEF error by horizon (left axis) and chunk-error violins (right axis).}
\label{fig:variation-by-source}
\end{figure}

\FloatBarrier

\subsection{Prediction Error Under Varying CFG}
\label{app:cfg-action-error}
\begin{figure}[ht]
\centering
\includegraphics[width=\linewidth]{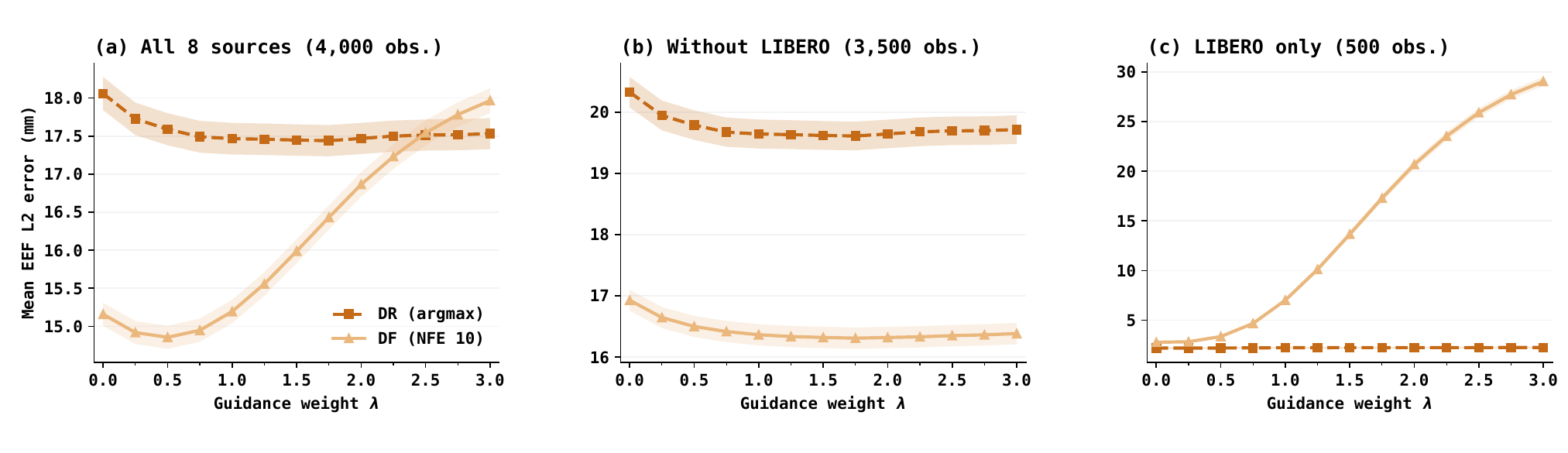}
\caption{\small \textbf{EEF error against ground truth as a function of the guidance weight $\lambda$.} Mean XYZ end-effector L2 error (mm) over the 24-step chunk, for DR (argmax decoding) and DF (NFE 10, mean over 32 draws). Shaded bands show $\pm1$ standard error over observations. $\lambda=0$ is unguided (\Cref{app:head-cfg}). \textbf{(a)} All 8 validation sources. \textbf{(b)} The 7 sources other than LIBERO. \textbf{(c)} LIBERO only.}
\label{fig:cfg-action-error}
\end{figure}

\Cref{tab:autoregression} reports that CFG raises DF's EEF error. To find the source of this increase, we sweep the guidance weight from $\lambda=0$ to $3$ in steps of $0.25$. We use 4{,}000 held-out observations, 500 from each of 8 validation sources (DROID, LIBERO, and six RoboCasa splits). The predictions come from the dense CFG sweep in \Cref{fig:token-ambiguity}(c). Error is the per-step XYZ Euclidean distance to the ground-truth action in physical units, averaged over the valid steps of the chunk, then over draws, then over observations. All heads are scored against the same target.

\confpar{Pooled over all sources (\Cref{fig:cfg-action-error}a).} DF has lower error than DR at small $\lambda$: 15.16\,mm versus 18.06\,mm unguided. DF's error is lowest at $\lambda=0.5$ (14.86\,mm) and then rises to 17.97\,mm at $\lambda=3$, passing DR at about $\lambda=2.5$. DR's error falls slightly with guidance and is nearly flat for $\lambda\ge1$ (17.44--17.53\,mm).

\subsection{State Responsiveness and Disturbance Stability}
\label{app:robustness}
\questionbox{Can a policy stay stable under small disturbances while still
responding to real state changes?}

We measure two properties of how a head's predictions change with the
observation. \emph{State responsiveness} asks whether the predicted chunk
changes as much as the demonstrated chunk when the state changes along a
real trajectory. \emph{Disturbance stability} asks how little the
distribution of predicted chunks moves after a small disturbance, compared
with the head's own sampling variation. The two pull against each other: a
head that changes little with its input is stable but under-responsive.
Flow heads have the highest disturbance stability, and within the flow
family DF is more responsive than CF. The state-responsiveness result in
\Cref{fig:state-responsiveness} is the lag-4 row of
\Cref{tab:expert-response}.

\subsubsection{Disturbance Stability}
\label{app:disturbance-stability}
Let $p_\pi(\cdot\mid o)$ be the distribution of policy $\pi$ over decoded, normalized
action chunks at observation $o$. From a simulator snapshot, executing the
nominal action for one control step gives observation $o$, and executing
the action perturbed by $\eta\sim\mathcal N(0,\epsilon^2I)$ in normalized
action coordinates gives $o'_\epsilon$. We draw two independent sets
$C_1,C_2$ of 32 chunks from $p_\pi(\cdot\mid o)$ and one set $P_2$ of 32
chunks from $p_\pi(\cdot\mid o'_\epsilon)$, with sampling noise independent
of $C_1$. With $\widehat W_2$ the empirical 2-Wasserstein distance between
two sets of chunks, computed by exact optimal assignment
(Equation~\eqref{eq:disturbance-w2}), we define
\begin{equation}
 F_\pi=\langle\widehat W_2(C_1,C_2)\rangle,\qquad
 T_\pi(\epsilon)=\langle\widehat W_2(C_1,P_2)\rangle,\qquad
 R_\pi(\epsilon)=\frac{T_\pi(\epsilon)-F_\pi}{F_\pi},
 \label{eq:relative-excess-transport}
\end{equation}
where brackets average over snapshots and perturbations. The floor $F_\pi$
is the distance between two samples at the same observation, so it
measures sampling variation alone. The excess $T_\pi(\epsilon)-F_\pi$ is the
extra distance caused by the disturbance. The \emph{relative excess
transport} $R_\pi(\epsilon)$ expresses this excess as a fraction of the
floor: $R_\pi=0.2$ means the predictions after the disturbance are 20\%
farther away than a fresh sample at the same observation. A head has
higher disturbance stability when $R_\pi$ is smaller. CR is deterministic,
so $F_\pi=0$ and $R_\pi$ is undefined; for CR we report only the excess.
\Cref{tab:relative-disturbance} reports these quantities for each head,
together with the RMS proposal spread $\sigma$. Protocol details are in
\Cref{app:robustness-protocol}.

\begin{table}[t]
\centering\small
\caption{Proposal spread and response to one-step simulator disturbances.
$\sigma$ is mean nominal RMS spread over the full bank. The independent
32-vs-32 transport statistics use the existing 24-snapshot subsample per
head and every other perturbation. Triples correspond to
$\epsilon=0.1,0.3,1.0$. Relative excess is computed from the displayed
rounded transport means. CR's deterministic law has zero floor, so its
relative excess is undefined. Spread and transport use different subsets.}
\label{tab:relative-disturbance}
\setlength{\tabcolsep}{4pt}
\begin{tabular}{lrrll}
\toprule
Head & $\sigma$ & $F_\pi$ & $T_\pi-F_\pi$ & $R_\pi$\\
\midrule
DF & 1.217 & 1.245 & 0.233 / 0.490 / 1.071 & 0.19 / 0.39 / 0.86\\
CF & 0.764 & 0.591 & 0.036 / 0.102 / 0.403 & 0.06 / 0.17 / 0.68\\
DR & 0.182 & 0.141 & 0.301 / 0.529 / 0.927 & 2.13 / 3.75 / 6.57\\
CR & 0.000 & 0.000 & 0.147 / 0.275 / 0.568 & undefined\\
\bottomrule
\end{tabular}
\end{table}

Flow heads have the highest disturbance stability. At $\epsilon=0.1$, the
excess transport is 19\% of the floor for DF and 6\% for CF, versus 213\%
for DR. Flow proposals are also broader than regression proposals, with
DF broader than CF ($\sigma=1.217$ versus $0.764$). DR's narrow proposal
distribution shifts substantially relative to its own variability; CR
also shifts despite having no proposal spread. This suggests a mechanism
by which regression errors could compound into out-of-distribution drift.

\subsubsection{State Responsiveness}
\label{app:state-responsiveness}
Take two frames $o_1,o_2$ of the same held-out demonstration, $k$ steps
apart (lag $k$). Let $\Delta s$ be the change in the normalized state,
$\Delta\ba^{\star}$ the change in the demonstrated chunk, and
$\Delta\mu_\pi=\mu_\pi(o_2)-\mu_\pi(o_1)$ the change in policy $\pi$'s mean
predicted chunk, where $\mu_\pi(o)=\E_{A\sim p_\pi(\cdot\mid o)}A$. Dividing a
chunk change by $\|\Delta s\|$ gives the empirical gain
$G_\pi=\|\Delta\ba_{1:H}\|/\|\Delta s\|$ shown in \Cref{fig:state-responsiveness}.
The head and the expert share $\Delta s$, so the ratio of their gains is
the ratio of their chunk changes. We define the state responsiveness of
policy $\pi$ at lag $k$ as
\begin{equation}
 \rho_\pi(k)=\frac{\operatorname{median}\|\Delta\mu_\pi\|}
 {\operatorname{median}\|\Delta\ba^{\star}\|},
 \label{eq:state-responsiveness}
\end{equation}
with medians over frame pairs. $\rho_\pi=1$ means the head changes as much
as the expert, and $\rho_\pi<1$ means it under-responds.
\Cref{tab:expert-response} reports the medians themselves; sampling
floors are not subtracted.

CF stays closer to nearby training actions than DF: its expert-calibrated
excess distance (Equation~\eqref{eq:expert-calibrated-excess}) is $-0.100$,
versus $-0.015$ for DF. This adherence can also make a policy too
``sticky.'' DF responds more than CF and is closer to the expert change at
every tested lag (\Cref{tab:expert-response}). At lag one, the expert
shift is 0.66, compared with 0.45 for DF and 0.36 for CF; at lag 16, the
values are 4.32, 3.55, and 3.28. At lag 4, $\rho_\pi$ is 0.94 for DR, 0.70
for DF, 0.63 for CF, and 0.58 for CR. Both flows under-respond, but CF does
so more strongly. DR tracks the expert most closely, with $\rho_\pi$ between
0.90 and 1.08 across lags, so state responsiveness and disturbance
stability are distinct requirements.

\begin{table}[ht]
\centering\small
\caption{Action changes on held-out expert trajectories. $\|\Delta s\|$
is the state displacement and expert $\Delta$chunk is the demonstrated
chunk change. Head columns report raw proposal-mean shifts; parentheses
show same-observation sampling floors for the stochastic heads. All values are medians over
pairs in normalized units; no floor subtraction or response-ratio
normalization is applied.}
\label{tab:expert-response}
\setlength{\tabcolsep}{4pt}
\begin{tabular}{rrrrrrr}
\toprule
Lag & $\|\Delta s\|$ & Expert $\Delta$chunk & DR (floor) & CR & DF (floor) & CF (floor)\\
\midrule
1  & 0.018 & 0.66 & 0.71 (0.05) & 0.35 & 0.45 (0.38) & 0.36 (0.21)\\
2  & 0.036 & 1.13 & 1.16 (0.05) & 0.62 & 0.77 (0.37) & 0.64 (0.22)\\
4  & 0.073 & 1.88 & 1.77 (0.05) & 1.09 & 1.32 (0.38) & 1.18 (0.21)\\
8  & 0.149 & 3.00 & 2.71 (0.05) & 1.93 & 2.34 (0.38) & 2.25 (0.22)\\
16 & 0.292 & 4.32 & 4.01 (0.05) & 3.13 & 3.55 (0.38) & 3.28 (0.22)\\
\bottomrule
\end{tabular}
\end{table}

\takeaway{Flow heads have the highest disturbance stability. Within the
flow family, CF stays closer to training actions but has lower state
responsiveness; DF keeps broader proposals and more of the expert's
response. This is the tradeoff between disturbance stability and state
responsiveness.}

\FloatBarrier

\subsection{Disturbance and Expert-Reference Diagnostic Protocol}
\label{app:robustness-protocol}
This appendix gives the measurement definitions and sampling protocol for
Appendix~\ref{app:robustness}. DR uses its deployed argmax decoder; its
residual proposal spread comes from varying random source-token inputs. CR is deterministic.

\subsubsection{Measuring a Distribution's Response to Disturbance}
Let $p_\pi(\cdot\mid o)$ be policy $\pi$'s law over decoded, normalized action
chunks. Its spread and its sensitivity are different quantities:
\begin{align}
 \sigma_\pi(o)&=\left(\E_{A\sim p_\pi(\cdot\mid o)}
       \|A-\mu_\pi(o)\|_2^2\right)^{1/2},\\
 W_2(p,q)&=\left[\inf_{\gamma\in\Pi(p,q)}
       \E_{(A,B)\sim\gamma}\|A-B\|_2^2\right]^{1/2},
 \label{eq:disturbance-w2}
\end{align}
where $\mu_\pi(o)=\E[A\mid o]$ and $\Pi(p,q)$ denotes joint distributions
with marginals $p,q$. Spread measures diversity at one observation;
$W_2(p_\pi(\cdot\mid o),p_\pi(\cdot\mid o'))$ measures how far probability
mass must move after a disturbance. Neither measures support cardinality.

We intervene on the simulator, because adding noise only to the proprioceptive input produced no measurable response. At 144 snapshots from DF rollouts across 18 RoboCasa atomic tasks,
we restore the simulator state and execute either the nominal action or an
action perturbed by $\eta\sim\mathcal N(0,\epsilon^2I)$ in normalized
action coordinates. After one control step, we render the resulting
observations and query each head at CFG~0 (NFE~10 for flows), drawing 64
chunks per observation.

\paragraph{Relative change requires a same-observation baseline.}
Even two independent finite samples from the same law have nonzero
empirical $W_2$, so $T_\pi$ alone would overstate the effect of a
disturbance. The floor $F_\pi$ in Equation~\eqref{eq:relative-excess-transport}
removes this baseline. Each of $C_1$, $C_2$, and $P_2$ contains 32 chunks,
and $R_\pi$ is a ratio of aggregate means. The ratio depends on sample count
and distribution shape; a large floor can reduce it. We therefore report
spread, absolute excess, and relative excess together.

\paragraph{Reference adherence must be calibrated against real actions.}
For held-out observation $o$, let $M(o)$ contain the $k=5$ nearest training
action chunks, with neighbors selected in normalized proprioception and
restricted to the same instruction when enough matches exist. Define
\begin{equation}
 r_0(A;o)=\min_{B\in M(o)}\|A-B\|_2,\qquad
 E_\pi=\frac{\E_o[\E_{A\sim p_\pi}r_0(A;o)-r_0(A_{\rm gt};o)]}
 {\operatorname{median}_o r_0(A_{\rm gt};o)}.
 \label{eq:expert-calibrated-excess}
\end{equation}
We compute $E_\pi$ within each source, then average over eight sources
(512 held-out observations per source, 16 proposals each).
The held-out expert reference matters: genuine demonstrated actions also
lie some distance from a finite training set. A negative excess means
predictions lie closer to this reference than held-out demonstrations do.

\paragraph{Gripper support.} Under a bin-scale tolerance, the fraction of gripper commands outside the demonstrated values is about 0.22 for discrete heads and 0.78--0.94 for continuous heads.

The raw on-data shifts in Table~\ref{tab:expert-response} compare change magnitudes. Adjacent chunks also shift their temporal window, so the comparison includes
replanning alignment. Sampling floors are reported separately and are not
subtracted from these shifts.

The one-step disturbance changes normalized proprioception by about
0.003--0.021, below the reference data's median neighbor spacing of 0.32.
The diagnostics use CFG~0 and
NFE~10, while reported control success uses separately selected settings.

%% file: appendix/factorization_error.tex
\section{Factorization Error Experiments}\label{app:factorization}

This section supplements \Cref{sec:factorization}.

\subsection{Experimental Methodology}\label{app:factorization-method}

\paragraph{Mode-detection criteria.}
For each translation-token distribution $p_j(b)$, we smooth probabilities
along the bin axis with a Gaussian kernel of standard deviation two bins,
normalized over the available bins at each location. We identify local
maxima, including endpoint maxima, in the smoothed distribution. Each bin
is assigned to its nearest peak, and each peak's mass is the sum of the
\emph{unsmoothed} probabilities assigned to it. We retain peaks with at
least 10\% mass. Two adjacent retained peaks qualify as separated modes
if they are at least eight bins apart and the minimum smoothed probability
between them is at most 50\% of the smaller peak height. Each translation token represents one coordinate at one predicted time step.
We call it ambiguous if it has a qualifying pair of modes, and count it
once regardless of how many such pairs it has. The mean count averages
ambiguous tokens per chunk. The threshold statistic reports the fraction
of chunks containing at least $k$ ambiguous tokens. For DF, detection is
performed on each final conditional token prediction before averaging over refinement draws.

\begin{figure}[!tb]
\centering
\includegraphics[width=\linewidth]{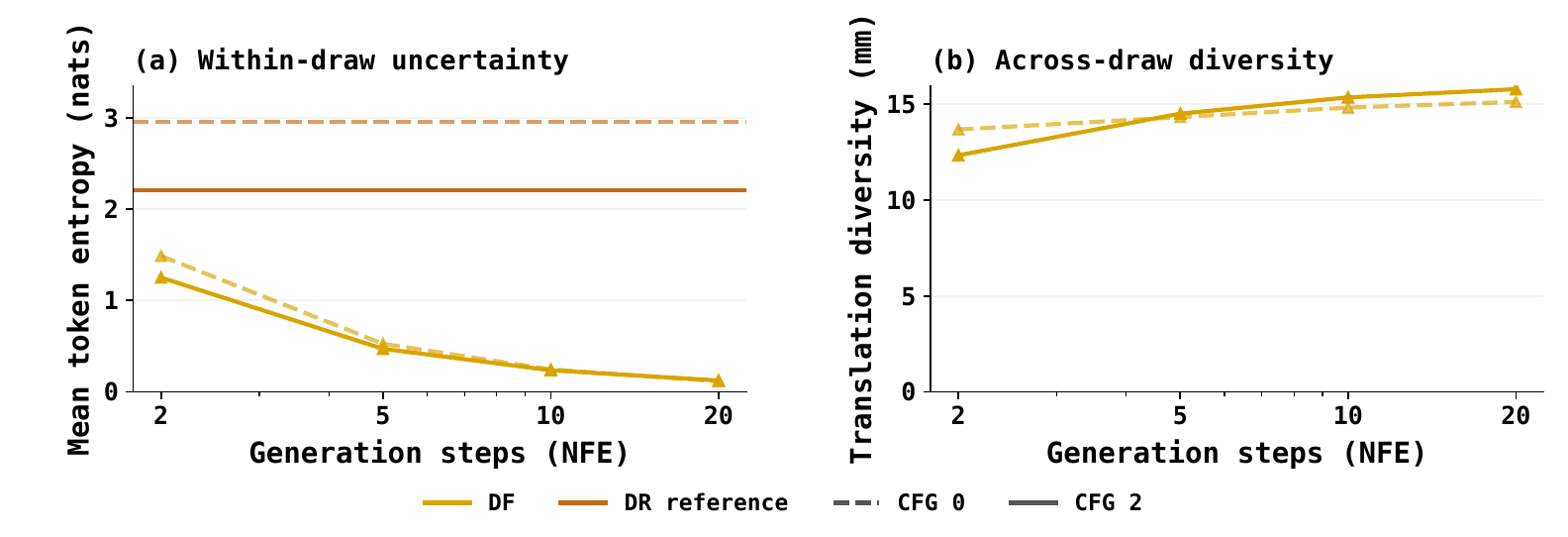}
\caption{\textbf{Refinement sharpens conditionals while retaining across-draw variation.}
Mean within-draw translation-token entropy (a) and across-draw
translation-command diversity (b) use the same 4,000 validation observations
and 16 draws per observation as Figure~\ref{fig:token-ambiguity}.
Gold curves show DF; orange horizontal lines are DR references.
Dashed and solid curves denote CFG~0 and CFG~2. From NFE~5 to~20,
entropy decreases while action diversity persists. Diversity is the RMS
pairwise Euclidean translation-command distance across draws, averaged
over horizon positions and observations, in mm.}
\label{fig:nfe-entropy-diversity}
\end{figure}

\paragraph{NFE sweep in Figure~\ref{fig:token-ambiguity}(a,b).}
We use 4,000 validation observations (500 from each of eight sources) and
16 DF draws per observation at each NFE and CFG~0. Panel (a) reports
the mean number of ambiguous tokens per chunk under the mode detector above;
Figure~\ref{fig:ambiguity-width}(a) gives the threshold curves.
Panel (b) reports percentiles of $\exp(\mathcal{H}(p))$ over all translation
tokens, draws, and observations. DR references use the same observations and
detector. Because the final query time is $t=1-1/\mathrm{NFE}$, the sweep
changes both the number of refinement steps and the time of the last
prediction. The mean count is 4.94 at NFE~2 and falls to 1.50, 0.57, and
0.22 at NFE~5, 10, and 20 (DR: 3.01).
Figure~\ref{fig:nfe-entropy-diversity} reports entropy and action
diversity for the same sweep.

\Needspace{14\baselineskip}
\subsection{Factorization Error and \texorpdfstring{$L_2$}{L2} Loss Diagnostic Results}\label{app:factorization-results}

\begin{figure}[!tb]
\centering
\begin{minipage}[t]{0.56\linewidth}
\centering
{\small (a) Ambiguous-token threshold curves}\\[2pt]
\includegraphics[width=\linewidth,trim=0 0 0 42bp,clip]{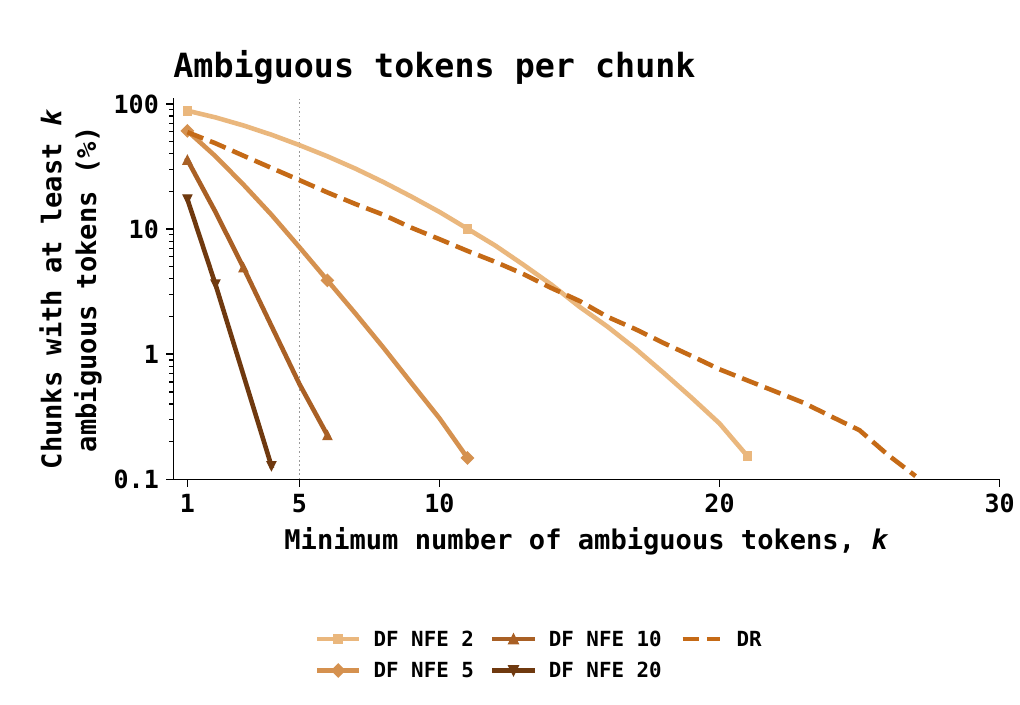}
\end{minipage}\hfill
\begin{minipage}[t]{0.40\linewidth}
\centering
{\small (b) Width by horizon}\\[2pt]
\includegraphics[width=0.78\linewidth,trim=0 0 0 50bp,clip]{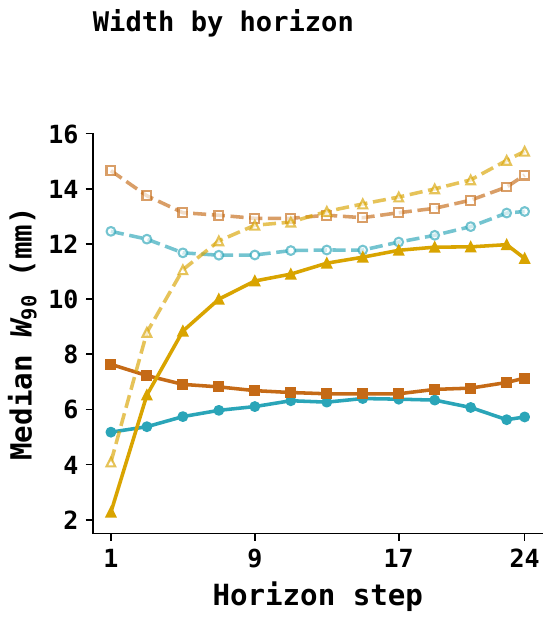}
\end{minipage}
\caption{\textbf{Ambiguity thresholds and width by horizon.}
(a) Percentage of chunks with at least $k$ ambiguous tokens at CFG~0, on the
same observations and draws as \Cref{fig:token-ambiguity}(a). The log axis stops
at 0.1\%.
(b) Median central-90\% width $W_{90}$ at each horizon step for CF (teal),
DR (orange), and DF (gold). Dashed lines with open markers are CFG~0; solid
lines with filled markers are CFG~2.}
\label{fig:ambiguity-width}
\end{figure}

\begin{wraptable}{r}{0.47\linewidth}
\vspace{-14pt}
\centering\small
\setlength{\tabcolsep}{3pt}
\caption{EEF target-prediction error (mm, $\downarrow$) on slow
validation near 100k steps, from single runs.}
\label{tab:eef-prediction-error}
\vspace{-6pt}
\footnotesize
\begin{tabular}{@{}lrrrr@{}}
\toprule
Source & CR & CF & DR & DF\\
\midrule
DROID & 21.60 & 26.60 & 25.09 & 19.67\\
LIBERO-plus & 9.88 & 7.23 & 2.53 & 2.90\\
\multicolumn{5}{@{}l}{RoboCasa}\\
\quad pretrain composite & 16.44 & 17.34 & 15.97 & 14.84\\
\quad MimicGen & 16.18 & 16.95 & 15.76 & 12.55\\
\quad target atomic & 17.34 & 18.97 & 19.55 & 15.62\\
\quad composite-seen & 12.11 & 14.87 & 15.39 & 12.93\\
\bottomrule
\end{tabular}
\vspace{-10pt}
\end{wraptable}
\Cref{tab:eef-prediction-error} reports the $L_2$
prediction errors referenced in \Cref{sec:discrete-vs-continuous}. CF
uses 100 integration steps and DF 10; regression heads use one pass. DF has
the lowest error on four of the six sources: DROID and three of the four
RoboCasa splits. DR has the lowest error on LIBERO-plus (2.53\,mm, against
9.88\,mm for CR), and CR on RoboCasa composite-seen. DF has lower error
than CF on all six sources. The remaining figures give additional
mode-ambiguity diagnostics for \Cref{fig:token-ambiguity};
\Cref{fig:ambiguity-width}(b) breaks $W_{90}$ down by horizon.

\begin{figure}[!tb]
\centering
\includegraphics[width=\linewidth]{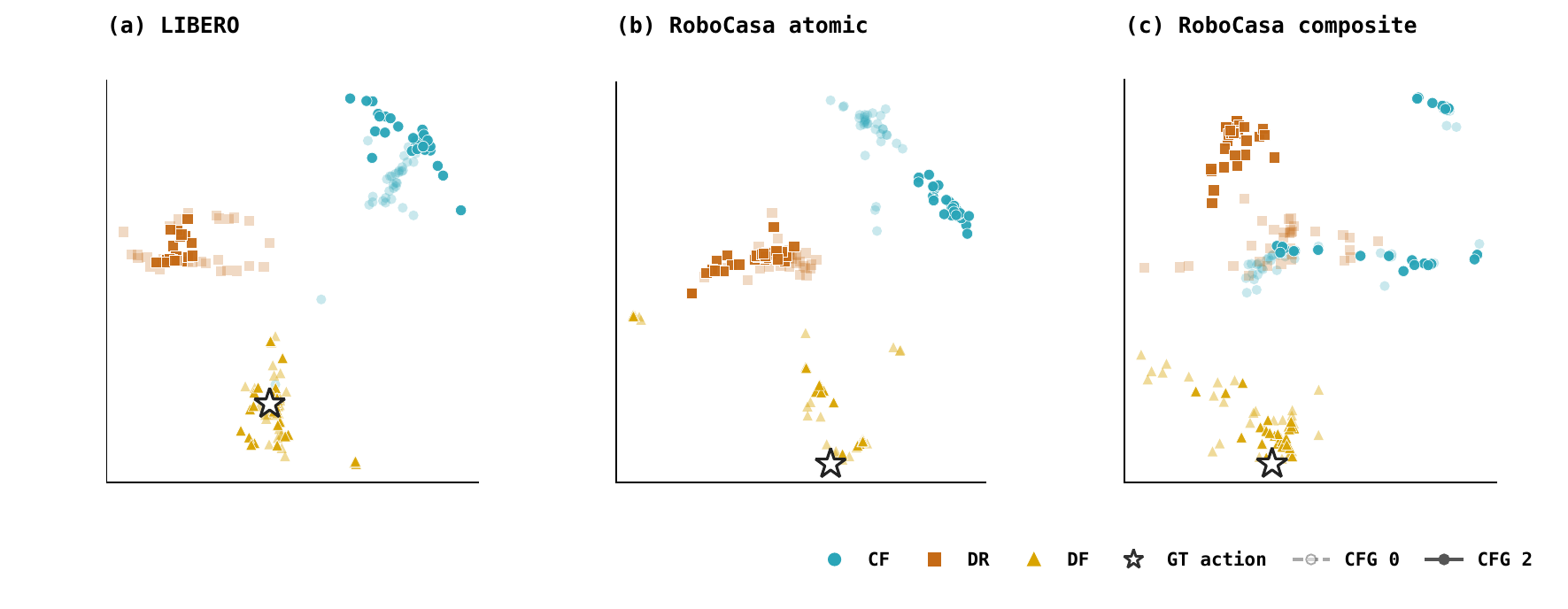}
\caption{\textbf{Additional t-SNE views of sampled action chunks.}
LIBERO, RoboCasa atomic, and RoboCasa composite supplement the DROID
example in Figure~\ref{fig:token-ambiguity}. Colors and marker shapes identify
CF, DR, and DF; faded and solid marks denote CFG~0 and CFG~2, respectively.
Stars denote reference actions. The original embeddings are preserved;
these projections provide a qualitative view of concentration under guidance.}
\label{fig:cfg-tsne}
\end{figure}

\subsection{Task Performance Under Varying NFE and CFG}\label{app:factorization-nfe-cfg}
We use the same evaluation protocol as in the rest of the paper
(\Cref{app:env-vla}), with one fixed checkpoint per head. Only the number of
decoding steps (NFE) and the guidance weight $\lambda$ (\Cref{app:head-cfg})
change.

\subsubsection{Task Performance Under Varying NFE}
\label{app:factorization-nfe}
\begin{figure}[!tb]
\centering
\includegraphics[width=\linewidth]{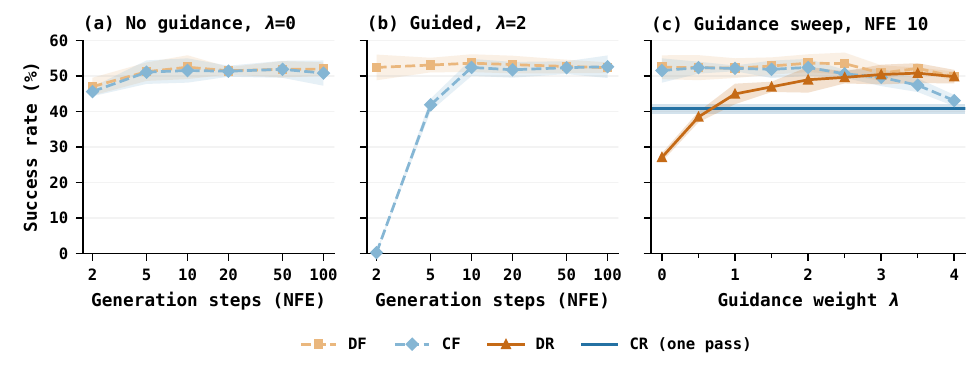}
\caption{\textbf{RoboCasa success rate under varying NFE and guidance.} Each
head uses one fixed checkpoint. (a,b) NFE sweep for DF and CF at guidance
weight $\lambda=0$ and $\lambda=2$. (c) Guidance sweep with DF and CF at
NFE~10; DR has no NFE axis. The CR line is an unguided one-pass reference.
Lines show the mean and bands the sample SD over three evaluation seeds.}
\label{fig:decoding-ablations}
\end{figure}

\Cref{fig:decoding-ablations}(a,b) shows that the effect of NFE depends on guidance.
Without guidance, both flow heads lose about five points at NFE~2, where DF
reaches $47.0\pm2.5\%$ and CF $45.6\pm1.3\%$. From NFE~5 to 100 they change
little: DF ranges from 51.2\% to 52.5\% and CF from 50.8\% to 51.9\%, with
no consistent gain from more steps. With $\lambda=2$, DF changes little at
every budget, ranging from 52.3\% to 53.6\%. Guided CF, however, fails under
coarse decoding: its success falls to $0.3\pm0.3\%$ at NFE~2 and
$41.9\pm1.7\%$ at NFE~5, then recovers to $52.4\pm2.4\%$ at NFE~10. Only
guided CF fails outright; unguided CF still reaches 45.6\% at NFE~2.

\subsubsection{Task Performance Under Varying CFG}
\label{app:factorization-cfg}
We next vary $\lambda$ with NFE fixed at 10 for both flow heads
(\Cref{fig:decoding-ablations}(c)); DR has no NFE axis. Guidance changes DR far
more than either flow head. DR improves from $27.1\pm1.2\%$ without guidance
to $48.9\pm3.6\%$ at $\lambda=2$, and reaches its highest mean,
$50.8\pm2.8\%$, at $\lambda=3.5$. Over the same range, its mean success on
composite-seen and composite-unseen tasks rises from about 10\% and 12\% to
34.8\% and 39.4\%.

Moderate guidance changes the flow heads only slightly: from $\lambda=0$ to
2, DF gains 1.1 points and CF 0.9 points. Strong guidance hurts, most of all
for CF, which falls to $43.1\pm1.6\%$ at $\lambda=4$, where DF obtains
$49.8\pm1.4\%$. The best guidance weight therefore depends on the head and
the decoding budget. For reference, CR reaches $40.8\pm1.4\%$ in one
unguided pass.

\FloatBarrier

\subsection{Visualization of Mode Inconsistency in RoboCasa}\label{app:factorization-robocasa}

\begin{figure}[!htb]
\centering
\includegraphics[width=0.88\linewidth]{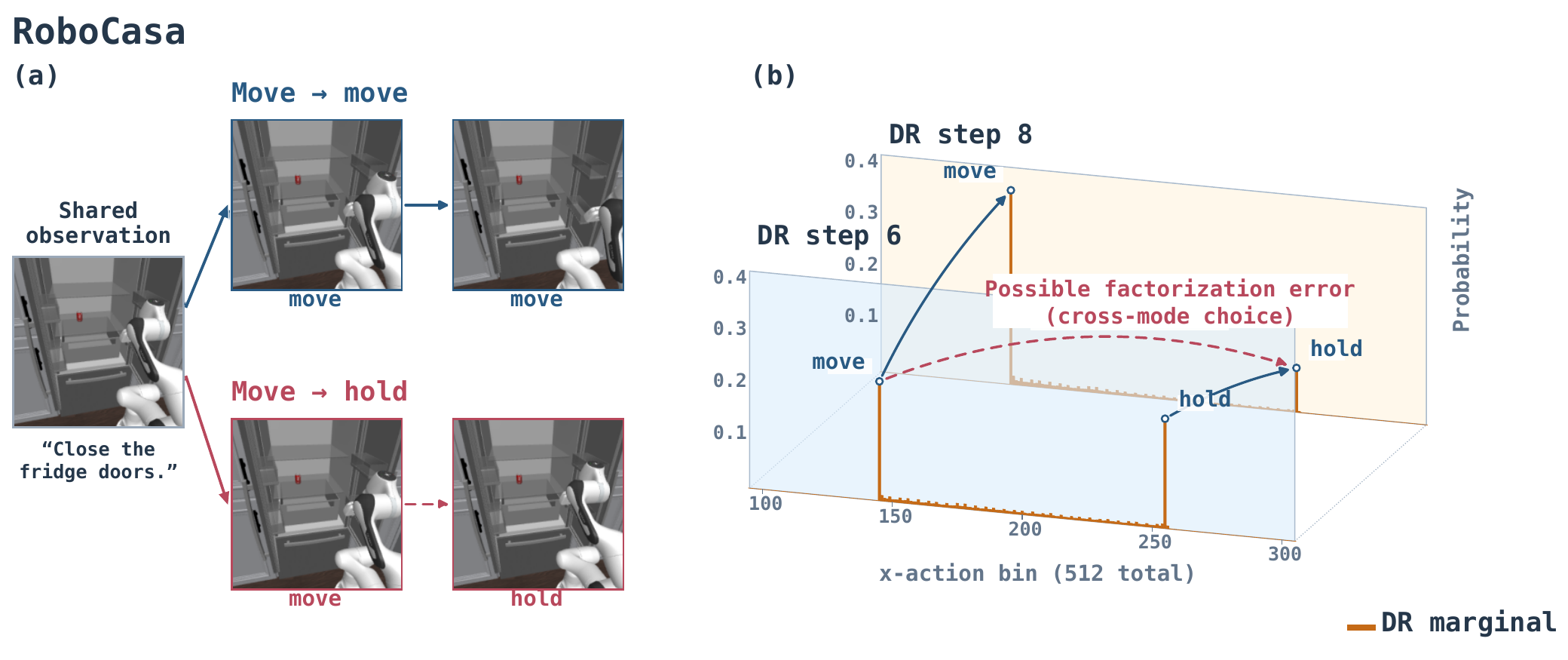}
\caption{\textbf{a} Robocasas example showing that in the same chunk, robot arm has two mode one is move consistenly, the second is first move then hold; \textbf{(b} We show that this independent token sampling could cause factorization error, leads to swithcing between motions on different timestep, the same as what we have seen in push-T ~\Cref{fig:pusht-factorization}}
\label{fig:robocasa-factorization}
\end{figure}

\paragraph{RoboCasa move--hold example.}
Figure~\ref{fig:robocasa-factorization} uses CloseFridge marginals and a
schematic mode illustration: moved thumbnails are taken from steps 12 and
24 of a DR-argmax replay; the hold thumbnail repeats the initial image.

%% file: appendix/tokenization_and_geometry.tex
\section{Tokenization and Geometric Priors}
\label{app:tokenization}
This section supplements \Cref{sec:geometry-and-tokenization}.
\Needspace{20\baselineskip}
\subsection{Performance Under Different Tokenization}
\label{app:tokenization-performance}
\begin{wraptable}{r}{0.43\linewidth}
\vspace{-14pt}
\centering\small
\setlength{\tabcolsep}{3pt}
\caption{\textbf{RoboCasa success by tokenizer and head.} Success rate
(\%) as mean $\pm$ standard deviation over five evaluation seeds, at the
best operating point for each tokenizer and head.}
\label{tab:tokenizer-control}
\vspace{-6pt}
\begin{tabular}{@{}llccc@{}}
\toprule
Tokenizer & Head & CFG & NFE & SR (\%)\\
\midrule
Frequency bins & DR & 0 & -- & $49.8 \pm 2.9$\\
               & DF & 2 & 20 & $46.0 \pm 1.6$\\
\quad + metric path & DF & 2 & 20 & $51.8 \pm 1.6$\\
\midrule
nFSQ & DR & 2 & -- & $41.6 \pm 3.2$\\
     & DF & 2 & 5  & $52.4 \pm 1.4$\\
\midrule
RVQ  & DR & 0 & -- & $14.4 \pm 1.3$\\
     & DF & 2 & 50$^*$ & $38.2 \pm 1.6$\\
\midrule
FAST & AR & 2 & -- & $48.2 \pm 2.2$\\
     & DF & 2 & 20 & $46.2 \pm 2.6$\\
\bottomrule
\end{tabular}\\[2pt]
{\footnotesize $^*$RVQ needs more NFE to decode, so DF RVQ uses NFE~50.}
\vspace{-10pt}
\end{wraptable}
\paragraph{Closed-loop results for tokenizers.}
Table~\ref{tab:tokenizer-control} compares nFSQ, RVQ, and FAST with their
frequency-bin baselines. For each tokenizer and
head, we report the operating point with the best pooled success, selecting
CFG, the replanning horizon, and, for DF, the NFE among 5, 10, and 20
(50 for RVQ).
The same setting is used on all three suites. Pooled success combines 180
atomic-seen episodes and 160 episodes from each composite suite.

The nFSQ result (52.4\%) is slightly above the frequency-bin metric-path
result (51.8\%). Neither exceeds the main mixture-path DF result of
53.9\%, which uses a different recipe and horizon.

\paragraph{FAST framing and validity.}
FAST compresses action sequences \citep{pertsch2025fast}, so generation
must respect its sequence grammar. During
training, valid decodes were near 100\% for AR, 25--40\% for DF, and 0\%
for DR. The FAST DR run still had 0\% validity at the end of training, so it
has no closed-loop evaluation. Flat scalar bins do not have this failure: every token sequence
decodes. With EOS-terminated framing, FAST AR reaches 48.2\% with
CFG~2; FAST DF reaches 46.2\%.

\paragraph{Frequency bins and DR.}
No tokenizer lifts DF above the best flat-bin DF, but frequency bins behave
differently for DR: CFG~0 and CFG~2 give almost the same success (49.8\%
and 49.0\%), and at CFG~0 DR is far above flat-bin DR (27.1\%,
\Cref{sec:factorization}). With frequency bins every token is used equally
often, so each token carries more information and should be harder to
predict, yet single-pass DR handles it well. We do not yet have a concrete
explanation, but the results suggest that when inference computation is
limited, how actions are encoded matters more.

\subsection{Geometry-Aware Probability Paths and Evaluations}
\label{app:metric-paths}
This appendix specifies the metric-path interventions summarized in
Section~\ref{sec:geometry-and-tokenization}. The path illustration is analytical; the
success rates below come from trained-policy evaluations.

\subsubsection{From a Mixture Path to a Metric-Induced Path}
Let $y\in\{0,\ldots,B-1\}$ denote a clean action token and let
$s\in[0,1]$ increase from noise toward the clean action. A probability path
specifies the distribution of corrupted tokens presented to the model at
each stage. With a uniform source, the mixture path is
\begin{equation}
 p_s^{\mathrm{mix}}(k\mid y)
 =\frac{1-s}{B}+s\,\mathbf{1}\{k=y\}.
 \label{eq:mixture-path}
\end{equation}
Here $s$ is the mixture weight; a separate schedule maps inference time to
$s$. Increasing $s$ raises the probability of the clean token, while every
incorrect bin retains the same probability. The corruption does not
distinguish a neighboring action value from a distant one.

The metric-induced path introduces this distinction through a Gibbs kernel:
\begin{equation}
 p_s^{\mathrm{metric}}(k\mid y)
 =\frac{\exp[-\beta(s)d_p(k,y)]}
 {\sum_{j=0}^{B-1}\exp[-\beta(s)d_p(j,y)]},
 \qquad d_p(k,y)=\frac{|k-y|^p}{(B-1)/2}.
 \label{eq:metric-path}
\end{equation}

\begin{figure}[tb]
\centering
\begin{minipage}[c]{0.42\linewidth}
\centering
\includegraphics[width=\linewidth]{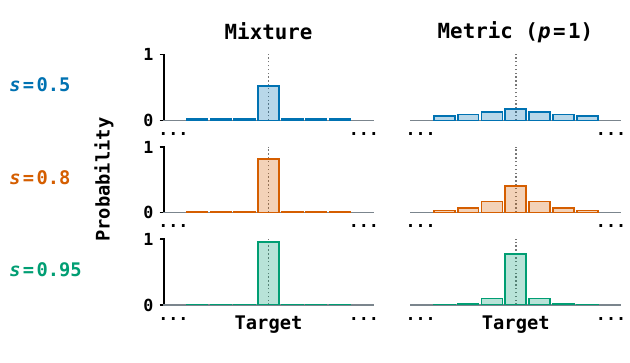}
\end{minipage}\hfill
\begin{minipage}[c]{0.54\linewidth}
\caption{Conditional corruption distributions at matched expected bin error.
Colors mark path progress $s=\alpha(t)$; the target is bin 10. Mixture
corruption keeps a uniform background, while metric corruption concentrates
around the target.}
\label{fig:metric-path-comparison}
\end{minipage}
\end{figure}
At $\beta=0$ the distribution is uniform; as $\beta$ increases, probability
concentrates around the clean bin, approaching a point mass as
$\beta\to\infty$. For finite positive $\beta$, nearby errors are more
probable than distant errors. We use $p=1$ for absolute bin distance and
$p=2$ for squared bin distance. Uniform scalar bins make index distance
proportional to distance in normalized action values; equal-frequency bins
instead impose this structure in quantile space.

Figure~\ref{fig:metric-path-comparison} isolates the difference at matched
expected bin error. The mixture path can achieve a low average error by
being exactly correct often, while its remaining errors are spread across
the vocabulary. The metric path instead allocates more of its error to
nearby bins. This changes the denoising problem even when the average
corruption magnitude is the same.

\paragraph{The path and its clock are separate choices.}
The Gibbs kernel defines a family of distributions; $\beta(s)$ determines
how quickly the model moves through that family. Write $p_\beta(k\mid y)$
for the kernel in Equation~\eqref{eq:metric-path} at inverse temperature
$\beta$, and let $\nu$ be a reference histogram over clean bins, uniform in
our runs. We use two path statistics, the expected absolute bin error and
the conditional entropy:
\begin{equation}
 \bar e(\beta)=\sum_{y}\nu(y)\sum_{k}p_\beta(k\mid y)\,|k-y|,
 \qquad
 \bar H(\beta)=\sum_{y}\nu(y)\,\mathcal{H}\bigl(p_\beta(\cdot\mid y)\bigr).
 \label{eq:metric-path-stats}
\end{equation}
The bin-progress clock makes $\bar e$ fall linearly in $s$, and the
entropy-progress clock does the same for $\bar H$:
\begin{equation}
 \bar e(\beta(s))=(1-s)\,\bar e(0)+s\,\bar e(\beta_{\max}),
 \qquad
 \bar H(\beta(s))=(1-s)\log B+s\,\bar H(\beta_{\max}).
 \label{eq:metric-bin-clock}
\end{equation}
At $\beta=0$ the kernel is uniform, so $\bar e(0)$ is the error of the
uniform source and $\bar H(0)=\log B$. The endpoint $\beta_{\max}=\beta(1)$
is the smallest $\beta$ at which every clean bin keeps its own token with
probability at least $1-q_{\mathrm{floor}}$, where
$q_{\mathrm{floor}}=1-(1-\rho)^{1/L}$ for $L$ tokens per chunk. We set
$\rho=0.05$, so at $s=1$ the chance that any token in the chunk is
corrupted is at most 5\%. Both clocks are numerically inverted and stored
as fixed lookup tables. Changing the clock therefore changes the training
and inference schedule without changing the form of the Gibbs kernel.

\paragraph{Learning clean tokens and decoding with distance-aware rates.}
Training retains clean-token cross-entropy,
$\E[-\log q_\theta(y\mid X_s,s,o)]$, summed over action-token positions;
$X_s$ is the corrupted chunk and $o$ is the observation. Geometry enters through the corruption path.
For a current token $i$ and a sampled clean endpoint $\hat y$ from the
network posterior, the implementation uses kinetic-optimal discrete flow
matching (KO-DFM) off-diagonal rates
\begin{equation}
 u_s(j\mid i,\hat y)
 =\dot\beta(s)\,p_s^{\mathrm{metric}}(j\mid\hat y)
   [d_p(i,\hat y)-d_p(j,\hat y)]_+,
 \qquad j\ne i.
 \label{eq:metric-rates}
\end{equation}
Here $[x]_+=\max(x,0)$. Conditional on that endpoint, only moves to closer bins have positive rates.
The decoder caps each token's rate-vector norm at 50 before the discrete
update.

\subsubsection{Evaluated Metric-Path Variants}
We trained two metric-path variants with RoboCasa evaluations: the
bin-progress clock with $p=1$ and the entropy-progress clock with $p=2$.
They differ in both the clock and the distance exponent. Their best pooled
success rates are 52.8\% and 4.6\%, respectively. The bin-progress clock
with $p=1$ on the equal-frequency representation reaches 51.8\%. Each rate uses one
operating point across all three suites. The first result is competitive
with, but does not exceed, the mixture-path DF result of 53.9\%.
The effect of explicit geometry depends strongly on how the path is
constructed.

\FloatBarrier

\subsection{Bin Representation and Input-Sensitivity Study}
\label{app:bin-geometry}
\begin{figure}[!tb]
\centering
\includegraphics[width=\linewidth]{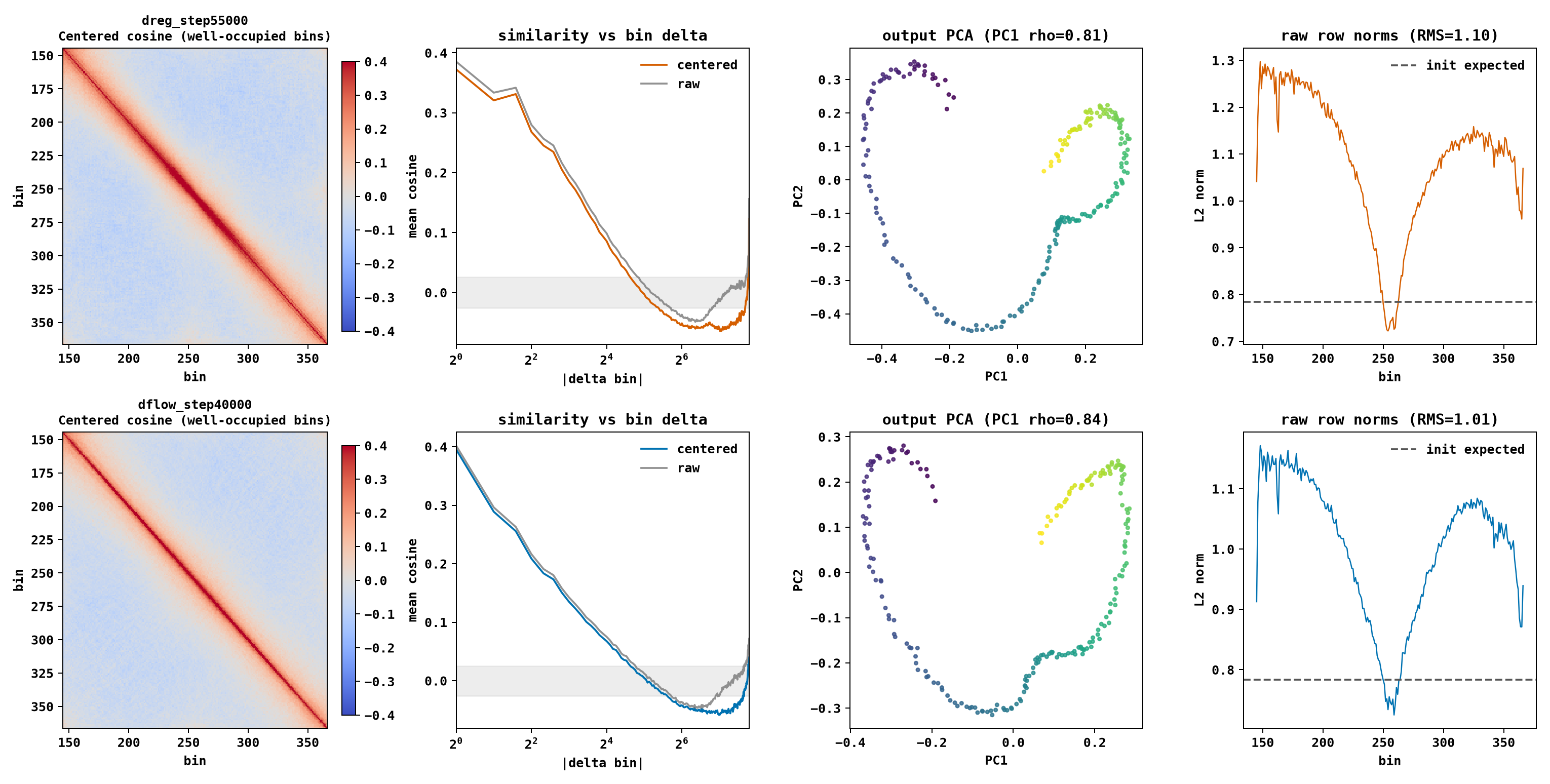}
\caption{Output-weight diagnostics: DR (top), DF (bottom).
Columns show centered cosine, lag-dependent similarity, PCA, and row norms.
The banded matrices and decreasing local similarity support ordinal
organization in both heads.
}
\end{figure}
\begin{figure}[!tb]
\centering
\includegraphics[width=\linewidth]{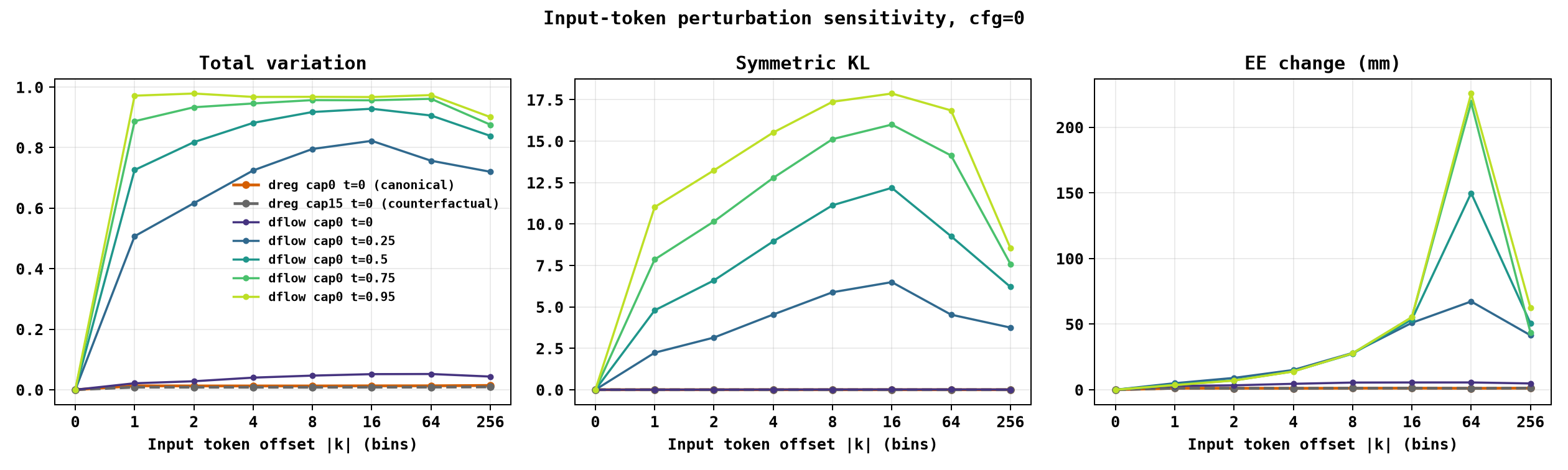}
\caption{Forward-pass input-offset probes. The plot also shows a counterfactual softcap-15 DR trace; our interpretation uses
only the canonical softcap-0 results. }
\end{figure}
We analyze DR and DF policies trained
on XDOF bus-table (plates) and bottle data. These policies use relative
stretch actions, a 16-step horizon, 14 action dimensions, a shared 512-bin
vocabulary, and 1,536-dimensional bin representations. The
checkpoints are DR step 55,000 and DF step 40,000.

Input tables subtract the global mean of
all vocabulary rows, while occupied output tables subtract the occupied-row
mean before cosine normalization. The analysis restricts
the headline statistics to bins with at least
50 aggregate occurrences in the validation sample: 222 bins (145--366).
Occupancy was counted over 100,794 valid frames, or 1,411,116 scalar action
values; 290 bins had no observed occupancy. The gate is important because
unused tails and common components can distort all-vocabulary similarities.
Bin separation is measured in the shared normalized coordinate system.
DR/DF centered input adjacency is $-0.0038/0.5733$, and output adjacency is
$0.3724/0.3953$. Output embedding-distance versus value-distance Spearman
correlations are $0.5835/0.5579$. Bin-relabeling permutation
tests give input adjacency $p=0.81/0.001$ and output adjacency $p=0.0005$
for both heads. DF input distances saturate
at larger separations (reported far-to-near slope ratio 0.027), so the result supports local order.
Positive timestep-embedding controls and shuffled-cluster negative controls
were also run. Multiple DR checkpoints retain near-zero
input adjacency.
Canonical posterior probes use 64 validation frames per head and report
14,252 valid output distributions per head; the input-sensitivity sweep
uses eight frames per head, with no classifier-free guidance and effective
logit softcap zero. DR total-variation responses remain approximately
0.0125--0.0141 over offsets of 1--256 bins. Replacing its random tokens with
true tokens changes decoded EEF position by 1.18\,mm. At DF time 0.75, offsets
of 1, 16, and 64 bins change the decoded normalized output by 0.0093, 0.1368,
and 0.5121. These interventions establish sensitivity to input tokens.

\subsubsection{Combined Output-Representation Visualization}
\Cref{fig:bin-geometry} combines these views. The output PCA projections
also organize bins by value, although their first
two components explain only about 11\% (DR) and 9\% (DF) of weight variance.
Output biases track the empirical marginal frequency of occupied bins (Pearson
$r=0.809/0.720$ for DR/DF with log-one-plus occupancy).

%% file: appendix/synthetic.tex
\section{Synthetic Fitting Protocol and Results}
\label{app:synthetic}
\subsection{Which Fine Structure Does Categorical Prediction Preserve?}
\label{app:synthetic-structure}
The robot experiments raise a simple question: can a model get the broad
motion right and still miss the fine detail? Robot data mixes too many
factors to answer this, so we use synthetic targets where we control the fine
structure directly.

\hypothesisheading{Hypothesis 2: categorical prediction can preserve steep fine structure.}
Following the multiscale construction of \citet{pan2025noising}, each target
is a broad sine plus a small fine-scale term. In the first target the fine
term is a fast sine. In the second it is a square wave, which jumps between
two values and so has no finite Lipschitz constant. CF and DF both predict
clean targets.

\begin{figure}[!htb]
\centering
\includegraphics[width=\linewidth]{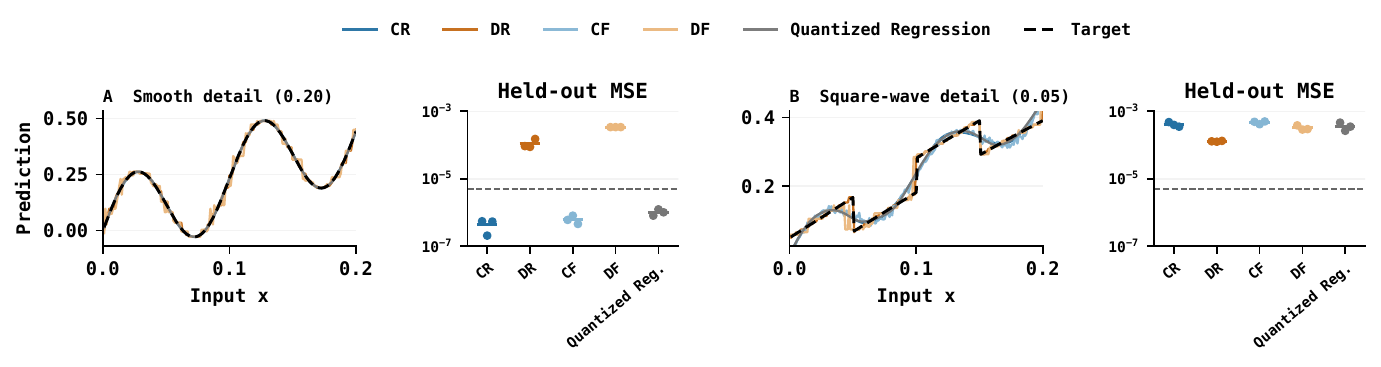}
\caption{Precision depends on target structure. Each target pairs original
predictions (seed 0) with held-out MSE for all three seeds; marks indicate
means and the dashed line indicates the hard-bin codec floor. Prediction
axes use the local target range; all errors remain included. CF/DF curves
show one draw, while stochastic MSE averages eight per-draw errors. Both
fine-term amplitude and shape change between targets.}
\label{fig:synthetic-regimes}
\end{figure}

\paragraph{Smooth fine detail: continuous regression wins.}
On the smooth target (Figure~\ref{fig:synthetic-regimes}A), CR's held-out
error is about 250 times lower than DR's. Both continuous heads go below the
codec's reconstruction floor. That floor limits any output decoded through
argmax, but not continuous outputs. The discrete heads stay well above it,
so bin resolution alone does not explain their larger error.

\paragraph{Steep fine detail: the ordering reverses.}
On the square-wave target (Figure~\ref{fig:synthetic-regimes}B), DR has the
lowest held-out error of all heads, with 3.2 times lower MSE and 4.5 times
lower MAE than CR. DF also beats both continuous heads, with 31 percent lower
error than CF. The prediction panel shows the difference. The regression
heads follow the steps but round off the corners, while the categorical heads
jump to the next bin right at the edge.

\paragraph{Controls narrow the explanation.}
Quantized Regression fits the bin centers with squared error. It behaves like
CR on both targets, so discrete target values are not what drives DR's
result. CF tracks CR and DF trails DR on both targets, so the reversal shows
up with one-pass and iterative prediction alike. That leaves the categorical
formulation as a whole: loss, output parameterization, and argmax decoding.
One likely reason is argmax itself. The logits can change smoothly while the
decoded bin jumps, which lets the model represent sharp transitions.